\def\year{2022}\relax
\documentclass[letterpaper]{article} 
\usepackage{aaai22}  
\usepackage{times}  
\usepackage{helvet}  
\usepackage{courier}  
\usepackage[hyphens]{url}  
\usepackage{graphicx} 
\usepackage{natbib}  
\usepackage{caption} 
\DeclareCaptionStyle{ruled}{labelfont=normalfont,labelsep=colon,strut=off} 
\usepackage{algorithm}
\usepackage{algorithmic}
\usepackage{booktabs}
\usepackage{multicol}
\usepackage{multirow}
\usepackage{xcolor}
\usepackage{amsmath}
\usepackage{amssymb}
\usepackage{pgfplots}
\usepackage{mathtools}
\usepackage{newfloat}
\usepackage{listings}
\floatstyle{ruled}
\newfloat{listing}{tb}{lst}{}
\floatname{listing}{Listing}
\usepackage[pagebackref=true,breaklinks=true,letterpaper=true,bookmarks=false]{hyperref}
\title{Learning Auxiliary Monocular Contexts Helps Monocular 3D Object Detection}
\author{
    Xianpeng Liu, \textsuperscript{\rm 1}
    Nan Xue, \textsuperscript{\rm 2} 
    Tianfu Wu \thanks{T. Wu is the corresponding author.} \textsuperscript{\rm 1}
}
\affiliations{
    \textsuperscript{\rm 1} Department of Electrical and Computer Engineering, North Carolina State University, USA\\
    \textsuperscript{\rm 2} School of Computer Science, Wuhan University, China \\
    {\tt xliu59@ncsu.edu, xuenan@whu.edu.cn, tianfu\_wu@ncsu.edu} \\
    {\fontsize{10}{10}\selectfont \url{https://git.io/MonoCon}}

}

\begin{document}

\maketitle

\begin{abstract}
Monocular 3D object detection aims to localize 3D bounding boxes in an input single 2D image. It is a highly challenging problem and remains open, especially when no extra information (e.g., depth, lidar and/or multi-frames) can be leveraged in training and/or inference.   
This paper proposes a simple yet effective formulation for monocular 3D object detection without exploiting any extra information. It presents the \textbf{MonoCon} method which learns \textbf{Mono}cular \textbf{Con}texts, as auxiliary tasks in training, to help monocular 3D object detection. 
\textit{The key idea is that with the annotated 3D bounding boxes of objects in an image, there is a rich set of well-posed projected 2D supervision signals available in training, such as the projected corner keypoints and their associated offset vectors with respect to the center of 2D bounding box, which should be exploited as auxiliary tasks in training.}
The proposed MonoCon is motivated by the Cram\`er–Wold theorem in measure theory at a high level.  
In implementation, it utilizes a very simple end-to-end design to justify the effectiveness of learning auxiliary monocular contexts, which  consists of three components: a Deep Neural Network (DNN) based feature backbone, a number of regression head branches for learning the essential parameters used in the 3D bounding box prediction, and a number of regression head branches for learning auxiliary contexts. After training, the auxiliary context regression branches are discarded for better inference efficiency. In experiments, the proposed MonoCon is tested in the KITTI benchmark (car, pedestrian and cyclist). It outperforms all prior arts in the leaderboard on the car category and obtains comparable performance on pedestrian and cyclist in terms of accuracy. Thanks to the simple design, the proposed MonoCon method obtains the fastest inference speed with 38.7 fps in comparisons.
\end{abstract}

\setlength{\abovecaptionskip}{1pt}
\setlength{\belowcaptionskip}{-6pt}
\setlength{\belowdisplayskip}{1pt} \setlength{\belowdisplayshortskip}{1pt}
\setlength{\abovedisplayskip}{1pt} \setlength{\abovedisplayshortskip}{1pt} 

\section{Introduction}

\begin{figure} [t]
    \centering
    \resizebox{0.99\linewidth}{!}{\input{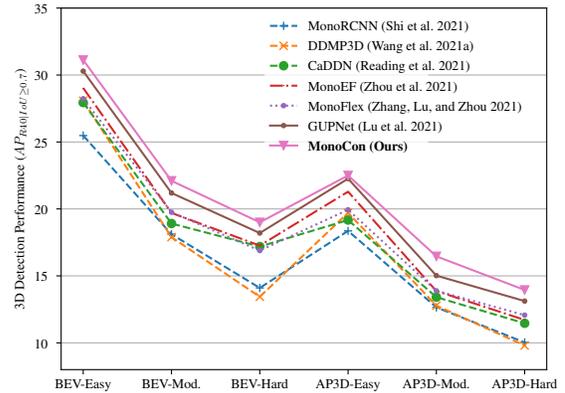}}
    \vspace{-2mm}
    \caption{\small Performance comparisons on the car category in the KITTI 3D object detection benchmark. The proposed MonoCon shows consistently better performance. See text for detail.}
    \label{fig:comparisons} \vspace{-5mm}
\end{figure}

3D object detection is a critical component in many computer vision applications in practice, such as autonomous driving and robot navigation. High performing methods often require more costly system setups such as Lidar sensors \cite{lidar3, lidar4, lidar1, lidar2} for precise depth measurements or stereo cameras \cite{s1, s2, s3, s4} for stereo depth estimation, and are often more computationally expensive. To alleviate those ``burden" and due to the potential prospects of reduced cost and increased modular redundancy, monocular 3D object detection that aims to localize 3D object bounding boxes from an input 2D image has emerged as a promising alternative approach with much attention received in the computer vision and AI community \cite{mono3d, roi10d, monodis, gs3d, m3drpn, rarnet, task3d, fda3d, dn3d, m3dssd, nms3d, fcos3d, pgd}. In addition to potential advantages in practice, developing powerful monocular 3D object detection systems will facilitate addressing a fundamental question in computer vision, that is whether it is possible to recover 3D structures from only 2D images which have lost the depth  information in the first place. 

\begin{figure*}[t]
    \centering
    \includegraphics[width=0.98\textwidth]{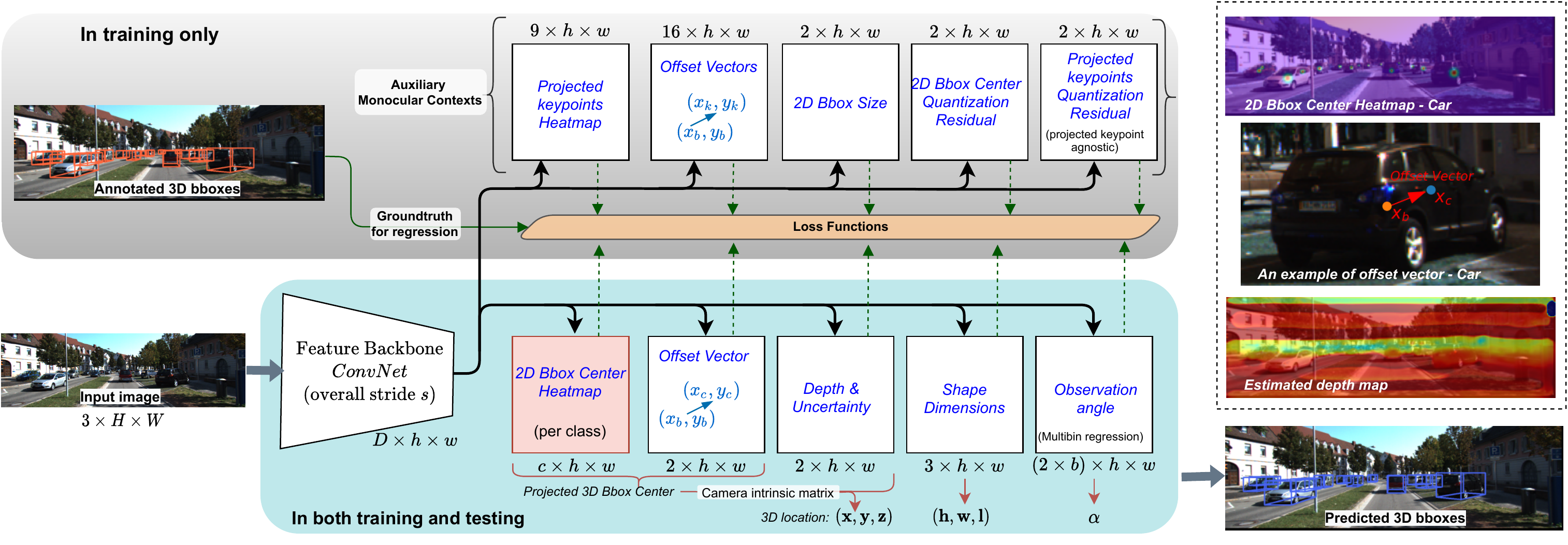}
    \caption{\small Illustration of the proposed MonoCon method for monocular 3D object detection without exploiting any extra information. It seeks a minimally-simple design. Given an input RGB image of dimensions $3\times H \times W$, a convolution neural network feature backbone computes the output feature map of dimensions $D\times h \times w$, where $D$ is the output feature map dimension, $h=H/s$ and $w=W/s$ with $s$ the overall stride/sub-sampling rate of the feature backbone (e.g., $s=4$). Then, light-weight regression head branches are used in a direct and straightforward way, including one set of the regression head branches for the essential parameters (3D locations, shape dimensions and  observation angles) which will be used in inferring the 3D bounding box, and the other set for the auxiliary contexts. Only the heatmap of 2D bounding box centers is class specific, and the others are class-agnostic. The proposed MonoCon is trained end-to-end and the auxiliary branches will be discarded in testing. In the right-top, the intermediate results for three regression branches are shown (note that the depth map will only be used sparsely based on the detected 2D bounding box centers). Best viewed in color and magnification. See text for detail. }
    \label{fig:workflow} 
    \vspace{-2mm}
\end{figure*}

This paper is interested in 3D object detection in the autonomous driving application. The objective is to estimate the 3D bounding box for each object instance such as a car in a 2D image. In the KITTI benchmark \cite{kitti}, a 3D bounding box is parameterized by: (i) the 3D center location $(\mathbf{x}, \mathbf{y}, \mathbf{z})$ in the camera 3D coordinates (in meters), (ii)  the observation angle $\alpha$ of the object  with respect to the camera based on the vector from the camera center to the 3D object center,  and (iii) the shape dimensions $(\mathbf{h}, \mathbf{w}, \mathbf{l})$, i.e., height, width and length (in meters).
Based on the extensive and insightful analyses made by the MonoDLE method~\cite{monodle} in the KITTI benchmark, 
\textbf{one main challenge of improving the overall performance in monocular 3D object detection lies in inferring the 3D center location with high accuracy}. 
  To address the challenge, there are two main types of settings in state-of-the-art monocular 3D object detection, depending on whether there are extra information (Lidar depth measurements,  monocular depth estimation results by a separately trained model or multi-frames) leveraged in training and/or  inference. In practice, the 3D center location $(\mathbf{x}, \mathbf{y}, \mathbf{z})$ is often decomposed to the projected 3D center in the image plane $(x_c, y_c)$ and the object depth $\mathbf{z}$. With the camera intrinsic matrix assumed to be known in both training and inference, the 3D location can be recovered with the inferred projected 3D center and object depth. 
  
  \textbf{This paper focuses on end-to-end monocular 3D object detection without exploiting any extra information.} 
It adopts the anchor-offset formulation proposed in the CenterNet~\cite{centernet} in learning the projected 3D center based on the 2D bounding box center (i.e., the anchor), and proposes a simple yet effective method that facilitates better overall performance (Fig.~\ref{fig:comparisons}). \textbf{The key idea is to leverage \textit{Mono}cular \textit{Con}texts as auxiliary learning tasks in training to improve the performance} (Fig.~\ref{fig:workflow}). The underlying rationale is that with the annotated 3D bounding boxes of objects in an image, there is a rich set of well-posed projected 2D supervision signals available in training, such as the projected corner keypoints and their associated offset vectors with respect to the anchor. They should be exploited in training to induce more expressive representations for monocular 3D object detection. The proposed method is thus dubbed as \textbf{MonoCon}. 

Statistically speaking, the monocular contexts can be treated as marginal random variables in the image plane, which are projected from the 3D bounding box random variables. In measure theory, the Cram\`er-Wold theorem~\cite{cramer1936some} states that a Borel probability measure on $\mathbb{R}^{k}$ is uniquely determined by the totality of its one-dimensional projections. Motivated by the Cram\`er-Wold theorem, the proposed MonoCon method introduces monocular projections as auxiliary tasks in training to learn more effective representations for monocular 3D object detection. In the meanwhile, it seeks a minimally-simple design of the overall detection system to justify the effectivenss of the underlying rationale and the high-level motivation.

In implementation, the proposed MonoCon utilizes a very simple design consisting of three components (Fig.~\ref{fig:workflow}): a Deep Neural Network (DNN) based feature backbone, a number of regression head branches for learning the essential parameters used in the 3D bounding box prediction, and a number of regression head branches for learning auxiliary contexts. After training, the auxiliary context regression branches are discarded. In experiments, the proposed MonoCon is tested in the KITTI benchmark (car, pedestrian and cyclist) \cite{kitti}. It outperforms prior arts (including methods that use lidar, depth or multi-frame extra information) in the leaderboard on the car category and obtains comparable performance on pedestrian and cyclist in terms of accuracy. Thanks to the simple design, the proposed MonoCon  obtains the fastest speed with 38.7 fps (on a single NVIDIA 2080Ti GPU card) in comparisons.

\section{Related Work and Our Contributions}

\textbf{Auxiliary tasks and auxiliary learning:} In machine learning, auxiliary tasks refer to tasks which are leveraged in training with the sole goal of better performing the primary tasks of interest in inference. The learning procedure is thus called auxiliary learning, in contrast to multitask learning, for which all tasks in training will be of interest in inference too. Auxiliary tasks and auxiliary learning have shown many successful applications in a wide range of fields including computer vision \cite{au1, au2, au6, au7, au8}, natural language processing \cite{au3} and reinforcement learning \cite{au4, au5}.  Although simple, exploiting comprehensive 2D auxiliary tasks has not been studied well in monocular 3D object detection. The proposed MonoCon showcases the advantage of auxiliary learning for both performance (on the car category) and inference efficiency in the KITTI benchmark~\cite{kitti}. 

\textbf{Monocular 3D detection \textit{with} extra information:} Monocular 3D detection falls behind lidar-based and stereo-image based counterparts significantly due to its ill-posed nature. Therefore, many monocular 3D detection methods seek solutions with the help of extra information, such as lidar data \cite{monorun, caddn}, off-the-shelf monocular depth estimation modules (pretrained using dense depth map) \cite{multifusion, d4lcn, pseudolidar, patchnet, am3d, ddmp3d}, multi-frames \cite{kinematic3d}, {or CAD models \cite{3dvp, deepmanta, mono3d++}}, etc. Although these methods have shown promising results, however, most of these models heavily rely on extra modules (i.e. depth estimation modules, etc.), which entails extra computation cost. As a result, these methods are usually slow in inference (less than 10 fps), severely hindering their applications in real-time autonomous driving. The proposed MonoCon method does not use any extra information and seeks a minimally-simple design with very promising performance and real-time inference speed achieved. One motivation is that before exploiting the more computationally expensive settings using multi-view images or more costly settings with more sensors, we want to understand the ``true" limit of purely monocular 3D detection methods.

\textbf{Monocular 3D detection \textit{without} extra information:} Since the seminal work of Deep3DBox \cite{deep3dbox}, many efforts \cite{fqnet, rtm3d, monods, monopair, object3d, monorcnn, ground-aware, monogeo, gupnet, monoflex} have been proposed to utilize 2D-3D geometric constraints to improve 3D detection performance, which are often posed as multi-task learning, rather than auxiliary learning. 
For example, in the RTM3D method~\cite{rtm3d}, all of the learned 2D tasks are used as optimization terms to calculate the 3D location of cars in the post-processing (using the PnP method) in inference.
In the MonoRCNN method~\cite{monorcnn}, one 2D task (i.e., 2D box) is used. The 2D box prediction is used to calculate the depth together with 3D box size in inference. The SMOKE method~\cite{smoke} does not learn any 2D task. One main claim in SMOKE is that 2D tasks will interfere the learning of 3D tasks.
In exploiting 2D-3D geometric constraints, existing work compute 3D locations explicitly based on 2D predictions, and thus often suffer from the well-known error amplification effect. So, more recent work try to use uncertainty modeling (e.g., the GUPNet \cite{gupnet}), or sophisticated model ensemble  (e.g., in MonoFlex \cite{monoflex}). The goal of the proposed MonoCon is to investigate the effects of 2D auxiliary tasks in training, and to eliminate the potential error amplification effect in inference, with improved performance obtained. 
More recently, there are efforts exploring how to generate extrinsic-invariant \cite{monoef} or distance-invariant \cite{movi3d} representations to improve 3D detection performance. The proposed MonoCon is complementary to aforementioned methods by leveraging well-posed 2D contexts projected from 3D bounding boxes as auxiliary learning tasks. It has the potential to be easily extended using aforementioned methods with performance further improved.

\textbf{Our contributions: } This paper makes three main contributions to monocular 3D object detection as follows: 
(i) It presents a simple yet surprisingly effective method, MonoCon for purely monocular 3D object detection by learning auxiliary monocular contexts. At a high level, the proposed MonoCon formulation can be explained by the Cram\`er-Wold theorem~\cite{cramer1936some} in measure theory. 
(ii) It shows state-of-the-art performance on the car category in the KITTI 3D object detection benchmark, outperforming prior arts by a large margin. It  obtains comparable performance on the pedestrian and cyclist categories. It can run at a speed of 38.7 fps, faster than prior arts. 
(iii) It sheds light on developing more powerful and efficient monocular 3D object detection systems by exploring and exploiting even more auxiliary contexts in general applications going beyond autonomous driving (e.g., robot navigation).

\section{Approach}

\subsection{Problem Definition}

Let $\Lambda$ be the image lattice (e.g. $384\times 1280$ in the KITTI benchmark), and $I_{\Lambda}$ an image defined on the lattice. As aforementioned, the objective of monocular 3D object detection is to infer the label (e.g., car, pedestrian and cyclist) and the 3D bounding box for each object instance in $I_{\Lambda}$. The 3D bounding box is parameterized by the 3D center location $(\mathbf{x}, \mathbf{y}, \mathbf{z})$ in meters, the shape dimensions $(\mathbf{h}, \mathbf{w}, \mathbf{l})$ in meters and the observation angle $\alpha \in [-\pi, \pi] $, all measured in the camera coordinate system. 
The observation angle is used in prediction due to its underlying stronger relationship with image appearance. The camera intrinsic matrix is assumed to be known in both training and inference. 

\textbf{Challenges.} Typically, both the shape dimensions and the orientation are directly regressed using features computed by a feature backbone such as a Convolutional Neural Network (CNN). The direct regression methods also have shown good performance for them individually. In the meanwhile, the overall 3D bounding box prediction performance (e.g., the Average Precision (AP) based on the intersection-over-union) is relatively less sensitive to the shape dimensions and the orientation, in the sense that if the 3D center location can be inferred with high accuracy, the AP will not decrease dramatically even if the shape dimensions and the orientation are not  accurately predicted. By contrast, even with very accurate estimate of shape dimensions and orientations, \textit{the AP will drop catastrophically if the 3D center location is perturbed}. The underlying reason is the significant gap between the shape dimensions (roughly between 1 and 3 meters) and the 3D location (roughly between 1 and 60 meters), and the uncertainty measured for both of them in monocular images will cause dramatically different effects for the overall AP.

There are two different formulations in learning the projected 3D center $(x_c, y_c)$: \textit{One} is to directly predict it by learning a heatmap representation, for which the projected centers falling outside the image plane are either simply discarded in training \cite{smoke, monodle} or cleverly handled with the help from the intersection point between the image edge and the line from the center of 2D bounding box to the outside projected 3D center \cite{monoflex}. \textit{The other} is to further decompose a projected 3D center into the center of 2D bounding box $(x_b, y_b)$ (i.e., the anchor inside the image plane) and an offset/displacement vector $(\Delta x, \Delta y)$ with $x_c=x_b+\Delta x$ and $y_c=y_b+\Delta y$, following the CenterNet \cite{centernet} formulation. Due to the large variation of the offset vectors, it is difficult to learn them. Thus, the latter is often inferior to the former in terms of the overall performance \cite{monoflex, monodle}, albeit it is an intuitively simple and generic representation for learning the projected 3D center. \textit{The proposed method shows that the latter can work well when sufficient monocular contexts are exploited in training. }

\subsection{The Proposed MonoCon Method}

As illustrated in Fig.~\ref{fig:workflow}, the proposed MonoCon method is simple by design, consisting of three components: 

\textbf{Feature Backbone.} Given an input RGB image $I_{\Lambda}$ of dimensions $3\times H \times W$, a feature backbone $f(\cdot;\Theta)$ is used to compute the output feature map $F$ of dimensions $D\times h \times w$, 
\begin{equation}
    F = f(I_{\Lambda}; \Theta),
\end{equation}
where $\Theta$ collects all the learnable parameters, $D$ is the output feature map dimension (e.g., $D=512$), and $h$ and $w$ are determined by the overall stride/sub-sampling rate $s$ in the backbone (e.g. $s=4$). We use the DLA network~\cite{dla} (DLA-34) that is widely used in monocular 3D object detection for fair comparisons in experiments. 

\textbf{The 3D Bounding Box Regression Heads.} We adopt the anchor-offset formulation in learning the projected 3D bounding box center $(x_c, y_c)$ in the image plane. {A regression head} is used to compute the class-specific heatmap $\mathcal{H}^{b}$ of dimensions $c\times h \times w$  for the 2D bounding box center $(x_b, y_b)$ for each of the $c$ classes (e.g., $c=3$ representing car, pedestrian and cyclist in the KITTI benchmark), 
\begin{equation}
    \mathcal{H}^{b} = g(F; \Phi^b), \label{eq:heatmap_b}
\end{equation}
where $g(\cdot; \Phi^b)$ is realized by a light-weight module with the learnable parameters $\Phi^b$, 
\begin{equation}
    F \xRightarrow[d\times 3\times 3 \times D]{Conv+AN+ReLU} \mathbb{F}_{d\times h \times w} \xRightarrow[c\times 1\times 1 \times d]{Conv} \mathcal{H}^{b}_{c\times h \times w}, \label{eq:regression}
\end{equation}
where the first convolution also reduce the feature dimension to $d$ (e.g., $d=64$) to be light-weight, and $AN$ represents the Attentive Normalization (AN)~\cite{an}, which is a light-weight module integrating feature normalization (e.g., BatchNorm~\cite{BatchNorm}) and channel-wise feature attention (e.g. the Squeeze-Excitation module~\cite{SE}).
Thanks to its mixture modeling formulation of the affine transformation in re-calibrating the features after standardization, it is adopted in the regression head for learning more expressive latent feature representations  $\mathbb{F}_{d\times h \times w}$. 
The light-weight module architecture $g(\cdot)$ (Eqn.~\ref{eq:regression}) is used by all regression heads with different instantiations (i.e., different learnable  parameters).  

The regression head of computing the offset vector $(\Delta x_b^c, \Delta y_b^c)$ from the 2D bounding box center $(x_b, y_b)$ to the projected 3D bounding box center $(x_c, y_c)$ is defined by, 
\begin{equation}
    \mathcal{O}^c_{2\times h \times w} = g(F;\Theta^{b_c}). \label{eq:offset_c}
\end{equation}

Similarly, the depth and the shape dimensions are regressed respectively as follows, 
\begin{align}
    \mathcal{Z}_{1\times h \times w} &= \frac{1}{\text{Sigmoid}\left(g\left(F; \Theta^Z\right)[0]\right)+\epsilon} - 1, \label{eq:depth}\\
    \sigma^{\mathcal{Z}}_{1\times h \times w} & = g(F; \Theta^Z)[1], \label{eq:depth_uncertainty} \\
    \mathcal{S}^{3D}_{3\times h \times w} &= g(F; \Theta^{S^{3D}}), \label{eq:shape_3D}
\end{align}
where $g(F; \Theta^Z)$ estimates the depth and its uncertainty. The inverse sigmoid transformation is applied to handle the unbounded output of $g(F; \Theta^Z)[0]$, as done in~\cite{centernet}, and $\epsilon$ is a small positive constant to ensure numeric stability. $\sigma^{\mathcal{Z}}$ is used to model the heteroscedastic aleatoric uncertainty in the depth estimation as done in~\cite{monopair, monoflex, monodle}. 

For the observation angle $\alpha$, the multi-bin setting proposed by~\cite{deep3dbox} is used. The angle range $[-\pi, \pi]$ is divided evenly into a predefined number of $b$ non-overlapping bins (e.g., $b=12$). The observation angle regression head is defined by, 
\begin{equation}
    \mathcal{A}_{2b\times h \times w} = g(F; \Theta^{A}), \label{eq:angle}
\end{equation}
where the observation angle $\alpha$ is predicted by computing its bin index, $\alpha_i\in \{0, 1, \cdots 11\}$ from the first $b$ channels (using $\arg\max$ after softmax along the $b$ channels) and the corresponding angle residual, $\alpha_r$ in the second $b$ channels of $\mathcal{A}$, together with proper conversions to ensure $\alpha\in [-\pi, \pi]$.

\textbf{Computing the predicted 3D bounding box.} Based on the peaks in each channel of the heatmap $\mathcal{H}^{b}$ (Eqn.~\ref{eq:heatmap_b}) after non-maximum suppression (NMS) and thresholding with a threshold $\tau$ (e.g., $\tau=0.2$), a set of 2D bounding box centers are detected for each class. Without loss of generality, consider a detected 2D bounding box center $(x_b, y_b)$ for a car, the offset vector is retrieved from $ \mathcal{O}^c$ (Eqn.~\ref{eq:offset_c}), $(\Delta x_b, \Delta y_b) =  \mathcal{O}^c(x_b, y_b)$. Then, the projected 3D center for the car is predicted by $(x_c, y_c)=(x_b+\Delta x_b, y_b + \Delta y_b)$. The corresponding depth is predicted by $z=\mathcal{Z}(x_b, y_b)$. With the camera intrinsic matrix, the 3D location $(\mathbf{x}, \mathbf{y}, \mathbf{z})$ will be computed in a straightforward way. Similarly, the shape dimensions $(\mathbf{h}, \mathbf{w}, \mathbf{l})$ and the observation angle $\alpha$ can be predicted for the car. With all these parameter inferred, the 3D bounding box is predicted.   

\textbf{The Auxiliary Context Regression Heads.} 
The proposed MonoCon method exploits four types of projection information from 3D bounding boxes as auxiliary learning tasks. 

\textit{i) The heatmaps of the projected keypoints.} As done in computing the 2D bounding box center heatmap $\mathcal{H}^b$ (Eqn.~\ref{eq:heatmap_b}), the first type of auxiliary contexts is the heatmaps of the 9 projected keypoints consisting of the projected 8 corner points and the projected center of the 3D bounding box, and we have, 
\begin{equation}
    \mathcal{H}^{k}_{9\times h \times w} = g(F; \Theta^k). \label{eq:heatmap_k}
\end{equation}

\textit{ii) The offset vectors for the 8 projected corner points.} In addition to the offset vector from the 2D bounding box center to the projected 3D bounding box center, $\mathcal{O}^c$ (Eqn.~\ref{eq:offset_c}), the second type of auxiliary contexts is the offset vectors from the 2D bounding box center to the 8 projected corner points of the 3D bounding box, and we have, 
\begin{equation}
    \mathcal{O}^{k}_{16\times h \times w} = g(F; \Theta^{b_k}). \label{eq:offset_k}
\end{equation}
Note that this is combined with Eqn.~\ref{eq:offset_c} with the first convolution block in $g(\cdot)$ shared in implementation. 

\textit{iii) The 2D bounding box size.} This is as done in the CenterNet~\cite{centernet}. The height and width of the 2D bounding box are regressed, 
\begin{equation}
    \mathcal{S}^{2D}_{2\times h\times w} = g(F;\Theta^{S^{2D}}). \label{eq:shape_2d}
\end{equation}

\textit{iv) The quantization residual of a keypoint location.} Due to the overall stride $s$ (typically $s>1$) in the feature backbone, there is a residual between the pixel location in the original input image $I_{\Lambda}$ and its corresponding  pixel location in the output feature map $F$ after multiplying the stride $s$. Consider the 2D bounding box center $(x^*_b, y^*_b)$ of a car in the original image, its pixel location in the feature map $F$ is $(x_b=\lfloor{\frac{x^*_b}{s}}\rfloor, y_b=\lfloor{\frac{y^*_b}{s}}\rfloor)$, and the residual is defined by, 
\begin{equation}
    \delta x_b = x^*_b - x_b, \quad \delta y_b = y^*_b - y_b. 
\end{equation}
We model the residual of the 2D bounding box center $(x_b, y_b)$ and that of the 9 projected keypoints $(x_k, y_k)$ separately to account for the underlying difference of the nature of those points. The latter is modeled in a keypoint-agnostic way as shown in Fig.~\ref{fig:workflow} for simplicity. We have, 
\begin{align}
    \mathcal{R}^{b}_{2\times h\times w} = g(F; \Theta^{R^b}), \label{eq:residual_b} \\
    \mathcal{R}^{k}_{2\times h\times w} = g(F; \Theta^{R^k}).  \label{eq:residual_k} 
\end{align}

\subsection{Loss Functions}
We use five loss functions which are widely used in monocular 3D object detection, consisting of (i) \textit{the Gaussian kernel weighted focal loss}~\cite{focalloss, cornernet} function for the heatmaps (Eqn.~\ref{eq:heatmap_b} and Eqn.~\ref{eq:heatmap_k}) as used in the CenterNet~\cite{centernet}, (ii) \textit{the Laplacian aleatoric uncertainty loss function }for the depth estimation (Eqn.~\ref{eq:depth} and Eqn.~\ref{eq:depth_uncertainty}), (iii)\textit{ the dimension-aware $L1$ loss function} for shape dimensions (Eqn.~\ref{eq:shape_3D}), (iv) \textit{the standard cross-entropy loss function} for the bin index in observation angles (Eqn.~\ref{eq:angle}), and (v)\textit{ the standard $L1$ loss function } for offset vectors (Eqn.~\ref{eq:offset_c} and Eqn.~\ref{eq:offset_k}),  the intra-bin angle residual in observation angles (Eqn.~\ref{eq:angle}), 2D bounding box sizes (Eqn.~\ref{eq:shape_2d}) and the quantization residual (Eqn.~\ref{eq:residual_b} and Eqn.~\ref{eq:residual_k}). We briefly discuss the first three as follows.

\textit{i) The Gaussian kernel weighted focal loss function for heatmaps}~\cite{focalloss, cornernet,centernet}.  Without loss of generality, consider a regressed heatmap $\mathcal{H}_{1\times h\times w}$ (e.g., the 2D bounding box centers of cars), the ground-truth heatmap $\mathcal{H}^*_{1\times h\times w}$ is also generated at the resolution of the regressed heatmap. For each ground-truth center point $(x^*_b, y^*_b)\in \mathbb{P}$ in the original image, its location in the ground-truth heatmap is $(x_b=\lfloor{\frac{x^*_b}{s}}\rfloor, y_b=\lfloor{\frac{y^*_b}{s}}\rfloor)$ (where $s$ is the overall stride of the feature backbone). A Gaussian kernel $G(x,y)=\exp{(-\frac{(x-x_b)^2+(y-y_b)^2}{2\cdot \sigma_b^2})}$ is used to model the center point, where $\sigma_b$ is a predefined object-size-adaptive standard deviation as used in~\cite{cornernet}. If two Gaussian kernels overlap, the element-wise maximum is kept. All the $G(\cdot, \cdot)$'s are then collapsed to form the ground-truth heatmap $\mathcal{H}^*$. The loss function is defined by, 
\begin{equation}
    \mathcal{L}(\mathcal{H}, \mathcal{H}^*)= \frac{-1}{N} \sum_{(x, y)} \begin{cases} (1-\mathcal{H}_{xy})^{\gamma}\log(\mathcal{H}_{xy}),  \text{ if } \mathcal{H}^*_{xy}=1, \\
    (1-\mathcal{H}^*_{xy})^{\beta}(\mathcal{H}_{xy})^{\gamma}\log(1-\mathcal{H}_{xy}),
    \end{cases}
\end{equation}
where $N=|\mathbb{P}|$ is the number of ground-truth points. $\beta$ and $\gamma$ are hyper-parameters (e.g., $\beta=4.0$ and $\gamma=2.0$).

\begin{table*}[t]
\begin{center}
\resizebox{0.94\textwidth}{!}{
    \begin{tabular}{l|c|c|ccc|ccc}
    \toprule
    \multirow{2}{*}{Methods, \textit{Publication Venues}} & \multirow{2}{*}{Extra Info.} & Runtime$\downarrow$ &
    \multicolumn{3}{c|}{\textit{$AP_{BEV|R40|IoU\ge0.7}\uparrow$}} & \multicolumn{3}{c}{\textit{$AP_{3D|R40|IoU\ge0.7}\uparrow$}} \\
     &  & (ms) & Easy & Mod. & Hard  & Easy & \textbf{Mod.} & Hard \\
    \midrule

    PatchNet, \textit{ECCV20} \cite{patchnet} & \multirow{3}{*}{Depth} & 400 & 22.97 & 16.86 & 14.97 & 15.68 & 11.12 & 10.17\\

    D4LCN, \textit{CVPR20} \cite{d4lcn} &  & 200 & 22.51 & 16.02 & 12.55 & 16.65 & 11.72 & 9.51\\
    
    DDMP-3D, \textit{CVPR21} \cite{ddmp3d} &  & 180 & 28.08 & 17.89 & 13.44 & 19.71 & 12.78 & 9.80\\
    
    \midrule
    
    Kinematic3D, \textit{ECCV20} \cite{kinematic3d} & Multi-frames & 120 & 26.69 & 17.52 & 13.10 & 19.07 & 12.72 & 9.17 \\
    
    \midrule

    MonoRUn, \textit{CVPR21} \cite{monorun} & \multirow{2}{*}{Lidar} & 70 & 27.94 & 17.34 & 15.24 & 19.65 & 12.30 & 10.58\\

    CaDDN, \textit{CVPR21} \cite{caddn} &  & 630 & 27.94 & 18.91 & 17.19 & 19.17 & 13.41 & 11.46 \\

    \midrule
    
    RTM3D, \textit{ECCV20} \cite{rtm3d} & \multirow{11}{*}{None} & 40 & 19.17 & 14.20 & 11.99 & 14.41 & 10.34 & 8.77\\
    
    Movi3D, \textit{ECCV20} \cite{movi3d} &   & 45 & 22.76 & 17.03 & 14.85 & 15.19 & 10.90 & 9.26\\
    
    IAFA, \textit{ACCV20} \cite{iafa} &   & 40 & 25.88 & 17.88 & 15.35 & 17.81 & 12.01 & 10.61\\

    MonoDLE, \textit{CVPR21} \cite{monodle} &   & 40 & 24.79 & 18.89 & 16.00 & 17.23 & 12.26 & 10.29\\

    MonoRCNN, \textit{ICCV21} \cite{monorcnn} &   & 70 & 25.48 & 18.11 & 14.10 & 18.36 & 12.65 & 10.03 \\
    
    Ground-Aware, \textit{RAL21} \cite{ground-aware} &   & 50 & 29.81 & 17.98 & 13.08 & 21.65 & 13.25 & 9.91 \\
    
    PCT, - \cite{pct} &   & 45 & 29.65 & 19.03 & 15.92 & 21.00 & 13.37 & 11.31 \\

    MonoGeo, - \cite{monogeo} & & 50 & 25.86 & 18.99 & 16.19 & 18.85 & 13.81 & 11.52 \\
    
    MonoEF, \textit{CVPR21} \cite{monoef} &   & 30 & 29.03 & 19.70 & 17.26 & 21.29 & 13.87 & 11.71\\

    MonoFlex, \textit{CVPR21} \cite{monoflex} &   & 35 & 28.23 & 19.75 & 16.89 & 19.94 & 13.89 & 12.07\\
    
    GUPNet, \textit{ICCV21} \cite{gupnet} & & 34 & \textcolor{blue}{30.29} & \textcolor{blue}{21.19} & \textcolor{blue}{18.20} & \textcolor{blue}{22.26} & \textcolor{blue}{15.02} & \textcolor{blue}{13.12} \\

    \midrule

    \multirow{5}{*}{\textbf{MonoCon (Ours)}, \textit{AAAI22}} & None & \textbf{25.8} & \textbf{31.12} & \textbf{22.10} & \textbf{19.00}
    & \textbf{22.50} & \textbf{16.46} & \textbf{13.95}\\
    
    \cmidrule{2-9}

     & \multirow{4}{*}{\textit{Improvement}} & v.s. Depth & +3.04 & +4.21 & +4.03
    & +2.79 & +3.68 & +3.78\\

     & & v.s. Multi-frames & +4.43 & +4.58 & +5.90
    & +3.43 & +3.74 & +4.78 \\

     & & v.s. LiDAR  & +3.18 & +3.19 & +1.81
    & +2.85 & +3.05 & +2.49\\

     & & v.s. None & +0.83 & +0.91 & +0.80
    & +0.24 & +1.44 & +0.83\\
    \bottomrule
    \end{tabular}}
\end{center}\vspace{-1mm}
\caption{Comparisons with state-of-the-art methods on the \textbf{car} category in the KITTI official \textit{\textbf{test}} set. Following the KITTI protocol, methods are ranked by their performance under the moderate difficulty setting. The best results are listed in \textbf{bold} and the second place in \textcolor{blue}{blue}.
The runtime of our MonoCon is measured using a single 2080Ti GPU card. The KITTI leaderboard entry of our MonoCon is \href{http://www.cvlibs.net/datasets/kitti/eval_object_detail.php?&result=59177e647b9ebb05bef36198cbaca61ff1d0eef2}{\textit{\textbf{at this link}}}. } \label{Tab:comparison_car} \vspace{-2mm}
\end{table*}

\textit{ii) The Laplacian aleatoric uncertainty loss function for depth}~\cite{monopair, monodle, monoflex}. 
Denote by $\mathcal{Z}^*_{1\times h\times w}$ the ground-truth (sparse) depth map in which the ground-truth depth of an annotated 3D bounding box is assigned to the corresponding ground-truth 2D bounding box center location in the lattice of $h\times w$, i.e., $\mathcal{Z}^*(x_b, y_b)$ (with the same inverse sigmoid transformation applied as in Eqn.~\ref{eq:depth}). The Laplace distribution is used in modeling the uncertainty $\sigma^Z$ (Eqn.~\ref{eq:depth_uncertainty}). For the prediction depth $\mathcal{Z}$ (Eqn.~\ref{eq:depth}), the loss function is defined by, 
\begin{equation}
    \mathcal{L}(\mathcal{Z}, \mathcal{Z}^*) = \frac{1}{|\mathbb{P}|} \sum_{(x_b, y_b)\in \mathbb{P}} \frac{\sqrt{2}}{\sigma^Z_b}|z_b - z^*_b| + \log (\sigma^Z_b), 
\end{equation}
where $\mathbb{P}$ is the set of ground-truth 2D bounding box center points, $\sigma^Z_b=\sigma^Z(x_b, y_b)$, $z_b=\mathcal{Z}(x_b, y_b)$ and $z^*_b=\mathcal{Z}^*(x_b, y_b)$.

\textit{iii) The dimension-aware $L1$ loss function for shape dimensions}~\cite{monodle}, which is motivated by the IoU oriented optimization~\cite{giou} and realizes a re-distribution of the standard $L1$ loss.  Similarly, let $\mathcal{S}^{3D^*}$ be the ground-truth map of shape dimensions assigned to the ground-truth 2D bounding box center locations in the lattice of $h\times w$. For the predicted shape dimensions $\mathcal{S}^{3D}$ (Eqn.~\ref{eq:shape_3D}), the loss function is defined by, 
\begin{equation}
    \mathcal{L}(\mathcal{S}^{3D}, \mathcal{S}^{3D^*}) = \lambda\cdot ||\frac{\mathcal{S}^{3D}- \mathcal{S}^{3D^*}}{\mathcal{S}^{3D}}||_1, 
\end{equation}
where $\lambda$ is the compensation weight to ensure the dimension-aware $L1$ loss has the same value as the standard $L1$ loss, which is by definition the ratio (without gradients in training) between the standard $L1$ loss and the dimension-aware loss before applying the compensation weight.  

\textbf{The Overall Loss} is simply the sum of all loss terms each of which has a trade-off weight parameter. For simplicity, we use 1.0 for all loss terms except for the 2D size $L1$ loss which uses 0.1. 

\section{Experiments}

In this section, we test the proposed MonoCon in the widely used and challenging KITTTI 3D object detection benchmark~\cite{kitti}. We first present comparisons with prior arts in the leaderboard, and then analyze the proposed MonoCon method using ablation studies.

\begin{table}[t]
\begin{center}
\resizebox{0.48\textwidth}{!}{
    \begin{tabular}{l|c|ccc|ccc}
    \toprule
    \multirow{2}{*}{Methods} & \multirow{2}{*}{Extra Info.} & \multicolumn{3}{c|}{\textit{Ped., $AP_{3D|R40|IoU\ge0.5}$}} & \multicolumn{3}{c}{\textit{Cyc., $AP_{3D|R40|IoU\ge0.5}$}} \\
     & & Easy & Mod. & Hard & Easy & Mod. & Hard \\
    \midrule
    
    DDMP-3D  & Depth & 4.93 & 2.55 & 3.01 & 4.18 & 2.50 & 2.32 \\
 
    CaDDN  & Lidar & 12.87 & 8.14 & 6.76 & \textbf{7.00} & \textbf{3.41} & \textbf{3.30}\\ 
    
    \midrule
    
    MonoDLE  & \multirow{6}{*}{None} & 9.64 & 6.55 & 5.44 & 4.59 & 2.66 & 2.45\\
    
    MonoGeo  & & 8.00 & 5.63 & 4.71 & 4.73 & 2.93 & 2.58 \\
    
    MonoEF  &  & 4.27 & 2.79 & 2.21 & 1.80 & 0.92 & 0.71\\
    
    MonoFlex  &  & 9.43 & 6.31 & 5.26 & 4.17 & 2.35 & 2.04\\
    
    GUPNet  & & \textbf{14.95} & \textbf{9.76} & \textbf{8.41} & \textcolor{blue}{5.58} & \textcolor{blue}{3.21} & \textcolor{blue}{2.66} \\

    \cline{3-8}

    \textbf{MonoCon (Ours)} &  & \textcolor{blue}{13.10} &\textcolor{blue}{8.41} & \textcolor{blue}{6.94} & 2.80 & 1.92 & 1.55 \\
    
    \bottomrule
    \end{tabular}
    }
\end{center}\vspace{-1mm}
\caption{Comparisons with state-of-the-art methods on the \textbf{pedestrian} category and the \textbf{cyclist} category in the KITTI official \textit{\textbf{test}} set.} \label{Tab:comparison_ped_cyc} \vspace{-4mm}
\end{table}

\textbf{Data.}  The KITTI dataset consists of 7,481 images for training and 7,518 images for testing. There are three categories of interest: car, pedestrian and cyclist.   The ground truth for the test set is reserved for evaluation on the test server.  In comparison with prior arts, we train our MonoCon on all 7,481 images and the performance is evaluated by the KITTI official server.  For ablation studies, we follow the protocol used by prior works~\cite{mono3d, chen2, multiview} to split the provided whole training data into a training subset (3,712 images) and a validation subset (3,769 images).  

\textbf{Evaluation Metrics.}  We follow the protocol provided in the KITTI benchmark.  The detection is evaluated by the average precision (AP) of 3D bounding boxes $AP_{3D|R40}$ and the AP of bird's eye view ($AP_{BEV|R40}$), both with $40$ recall positions ($R40$) used and under three difficulty settings (easy, moderate, and hard).\textit{ The moderate difficulty level is used to rank methods in the KITTI leaderboard.} The APs are computed with the intersection-over-union (IoU) threshold 0.7, 0.5 and 0.5 for car, pedestrian and cyclist respectively. 

\textbf{Implementation Details.} Our MonoCon is trained on a single GPU with a batch size of $8$ in an end-to-end way for 200 epochs. The AdamW optimizer is used with $(\beta_1, \beta_2)=(0.95, 0.99)$ and weight decay $0.00001$ (not applying to feature normalization layers and bias parameters). The initial learning rate is $2.25e-4$, and the cyclic learning rate scheduler is used (1 cycle), which first gradually increases the learning rate to $2.25e-3$ with the step ratio $0.4$, and then gradually drops to $2.25e-4\times 1.0e-4$ (i.e., the target ratio is $(10, 1.0e-4)$). The cyclic scheduler is also applied for the momentum with the target ratio $(0.85 / 0.95, 1)$ and the same step ratio $0.4$. Due to the auxiliary context regression heads in training, the complexity of training our MonoCon is greater in terms of training time and memory footprint. After training, the auxiliary components will be discarded, resulting in faster speed in inference than prior arts. {We adopt the commonly used data augmentation methods such as photometric distortion and random horizontal flipping following ~\cite{centernet, monodle, monoflex}. We also utilize random shifting to augment cropped instances at the edge of images.}

\subsection{Comparisons with State-of-the-Art Methods}
Table~\ref{Tab:comparison_car} and Table~\ref{Tab:comparison_ped_cyc} show the quantitative comparisons of our MonoCon with state-of-the-art methods. \textit{We provide qualitative results in the supplementary materials}.

\textbf{Comparisons on the car category.} Cars are the dominant objects in the KITTI 3D object detection benchmark, and of the most interest in evaluation. Our MonoCon consistently outperforms all prior arts. In terms of the KITTI ranking protocol based on the $AP_{3D|R40}$ under the moderate difficulty setting, our MonoCon achieves significant improvement by \textbf{$1.44\%$ absolute increase} against the runner-up method, the GUPNet~\cite{gupnet}. It also runs faster than prior arts. The improvement justify the effectiveness of learning more auxiliary monocular contexts in monocular 3D object detection for the autonomous driving applications. \textit{Our MonoCon also consistently outperform prior arts in the validation set with the results provided in the supplementary materials} due to the space limit. 

\textbf{Comparisons on the pedestrian and cyclist categories.} Our MonoCon shows inferior performance than some of the prior arts. On the pedestrian category, our MonoCon shows $1.35\%$ drop of $AP_{3D|R40}$ under the moderate setting comparing with the best model, the GUPNet~\cite{gupnet}, but outperforms all other methods in comparisons. On the cyclist category, our MonoCon shows $1.29\%$ drop comparing with the best purely monocular model, the MonoDLE method~\cite{monodle}. Overall, our MonoCon is less effective on the cyclist category among the three categories. We observe that the 3D bounding boxes of pedestrian and cyclist are much smaller than those of car, and the projected monocular contexts are often in the very close proximity in the feature map (of the $h\times w$ lattice). The close proximity may affect the learnability and effectiveness of the auxiliary contexts. One potential solution is to randomly sample a subset of auxiliary contexts that are spatially separate from each other, which we leave to future work.

\subsection{Ablation Studies}

We investigate the effects of learning auxiliary contexts and of the class-agnostic settings for regression heads in Fig.~\ref{fig:workflow}. 

\textbf{The Importance of Learning Auxiliary Contexts and the Attentive Normalization (AN).} Table~\ref{Tab:ablation_auxiliary} shows the comparisons which show the effectiveness of the proposed MonoCon and justify the importance of the design choices.  One the one hand, without the auxiliary components, our MonoCon is most similar to the MonoDLE method~\cite{monodle}. We retrain an enhanced  MonoDLE model which obtains significantly better performance than the vanilla MonoDLE. Our MonoCon still outperforms the enhanced MonoDLE by a large margin.  On the other hand, we test 7 variants of our MonoCon model from (a) to (g). The auxiliary contexts are significantly more important than the Attentive Normalization: (g) vs (a) with a 8.49\% absolution increase under the moderate difficulty settings, which clearly shows the significance of the proposed formulation against some implementation tuning. From (b) to (f), we rank the importance of the auxiliary contexts based on the performance of the model trained without them: the lower the performance is, the more important the context(s) are. 
\begin{table}[!h]
\begin{center}
\resizebox{0.48\textwidth}{!}{
    \begin{tabular}{c|ccccc|c|ccc}
    \toprule
     &\multicolumn{5}{c|}{\textit{2D Context Heads}} & \multirow{2}{*}{AN} & \multicolumn{3}{c}{\textit{Val, $AP_{R40}$, Car}}\\
      & $\mathcal{O}^{k}$ & $\mathcal{R}^{b}$ & $\mathcal{S}^{2D}$ & $\mathcal{H}^{k}$ & $\mathcal{R}^{k}$ & & Easy & Mod. & Hard\\

    \midrule
    MonoDLE & - & -  & \checkmark & - & - & - & 17.45 & 13.66 & 11.68 \\
    
    MonoDLE$^*$ & - & -  & \checkmark & - & - & \checkmark & 22.96 & 16.76 & 14.85 \\
    
    \midrule

    (a) & - & -  & - & - & - & \checkmark & 16.57 & 10.20 & 8.14 \\
    
    (b) & - & \checkmark & \checkmark & \checkmark & \checkmark & \checkmark & 17.65 & 11.42 & 8.71  \\
    
    (c) & \checkmark & \checkmark & \checkmark & - & - & \checkmark & 21.76 & 16.09 & 13.34 \\

    (d) & \checkmark & \checkmark & - & \checkmark & \checkmark & \checkmark & 22.72 & 16.68 & 13.88 \\
    
    (e) & \checkmark & \checkmark & \checkmark & \checkmark & - & \checkmark & 23.51  & 17.76 & 15.03 \\
    
    (f) & \checkmark & - & \checkmark & \checkmark & \checkmark & \checkmark & 24.11 & 18.28 & 15.35 \\
    
    (g) & \checkmark & \checkmark & \checkmark & \checkmark & \checkmark & - & 25.37 & 18.69 & 15.67 \\
    \midrule
    
    \textbf{MonoCon (Ours)} & \checkmark & \checkmark & \checkmark & \checkmark & \checkmark & \checkmark & \textbf{26.33} & \textbf{19.01} & \textbf{15.98} \\
    
    \bottomrule
    \end{tabular}
    }
\end{center}
\caption{\small Ablation studies on different auxiliary contexts (see Fig.~\ref{fig:workflow} and Eqn.~\ref{eq:heatmap_k} to Eqn.~\ref{eq:residual_k}) and the Attentive Normalization~\cite{an} (AN in Eqn.~\ref{eq:regression}) in our MonoCon. $^*$For fair comparisons and to justify our method's effectiveness, we re-implement and train a modified and enhanced version of the vanilla MonoDLE~\cite{monodle} with the AN added and the exactly same training settings as our MonoCon. \textit{Note that the 2D size is used in both training and inference in MononDLE.} }  \label{Tab:ablation_auxiliary} \vspace{-3mm}
\end{table}

\textbf{The effects of class-agnostic settings in regression heads and of training settings.} Table~\ref{Tab:ablation_regression_head} shows the comparisons. On the one hand, using the class-agnostic design shows better performance for the car and cyclist categories, while the class-specific design is significantly better for the pedestrian category. On the other hand, jointly training the three categories is beneficial, which indicates that some inter-category synergy may exist.
\begin{table}[!h]
\begin{center}
\resizebox{0.48\textwidth}{!}{
    \begin{tabular}{ccc|c|ccc|ccc|ccc}
    \toprule
    \multicolumn{3}{c|}{\textit{Training Data}} & \multirow{2}{*}{Class-Agnostic} & \multicolumn{3}{c|}{\textit{Val, $AP_{R40}$, Car}} & \multicolumn{3}{c|}{\textit{Val, $AP_{R40}$, Ped.}} & \multicolumn{3}{c}{\textit{Val, $AP_{R40}$, Cyc.}}  \\
    Car & Ped. & Cyc. & & Easy & Mod. & Hard  & Easy & Mod. & Hard & Easy & Mod. & Hard \\
    \midrule

    \checkmark & \checkmark & \checkmark & \checkmark & \textbf{26.33} & \textbf{19.01} & \textbf{15.98} & 1.46 & 1.31 & 0.99 & \textbf{7.60} & \textbf{4.35} & \textbf{3.55} \\
    
    \checkmark & \checkmark & \checkmark & - & 24.69 & 18.53 & 15.49 & \textbf{9.21} & \textbf{6.85} & \textbf{5.49} & {3.44} & 1.50 & 1.50 \\
    
    \midrule

    \checkmark & - & - & N/A & 24.60 & 18.15 & 15.36 & - & - & - & - & - & - \\

    - & \checkmark & - & N/A & - & - & - & 5.10 & 4.13 & 3.10 & - & - & - \\
    
    - & - & \checkmark & N/A & - & - & - & - & - & - & 2.98 & 1.66 & 1.27 \\
    
    \bottomrule
    \end{tabular}
    }
\end{center}
\caption{\small Ablation studies on the design of regression heads (class-agnostic vs class-specific in Eqn.~\ref{eq:offset_c} to Eqn.~\ref{eq:residual_k}) and the training settings (joint vs separate training of car, pedestrian and cyclist).} \label{Tab:ablation_regression_head} \vspace{-3mm}
\end{table}

\section{Conclusion}
This paper proposes a simple yet effective formulation for monocular 3D object detection without exploiting any extra information. It presents the MonoCon method which learns auxiliary monocular contexts that are projected from the 3D bounding boxes in training.  The proposed MonoCon utilizes a simple design in implementation consisting of a ConvNet feature backbone and a list of regression heads with the same module architecture for the essential parameters and the auxiliary contexts. In experiments, the proposed MonoCon is tested in the KITTI 3D object detection benchmark with state-of-the-art performance on the car category and comparable performance on the pedestrian and cyclist categories. At a high level, the effectiveness of the proposed MonoCon can be explained by the  Cram\`er–Wold theorem in measure theory. Ablation studies are performed to investigate, and the results support, the effectiveness of the proposed MonoCon method.

\section*{Acknowledgements}
X. Liu and T. Wu were supported in part by NSF IIS-1909644, ARO Grant W911NF1810295, NSF IIS-1822477, NSF CMMI-2024688, NSF IUSE-2013451 and DHHS-ACL Grant 90IFDV0017-01-00. N. Xue was supported  by NSFC under Grant 62101390.
The views presented in this paper are those of the authors and should not be interpreted as representing any funding agencies.

\bibliography{main}

\clearpage

\onecolumn
\section{Supplementary Materials}

\section{Quantitative and Qualitative Results on KITTI Validation Set}

\paragraph{Quantitative Results:} We present our quantitative 3D detection results on KITTI validation set ~\cite{mono3d} in Tab.~\ref{Tab:val}. It shows that our MonoCon method achieves significantly improvements on validation set for the car category.

\begin{table*}[ht]
\begin{center}

\begin{tabular}{l|c|ccc|ccc}
\toprule
\multirow{2}{*}{Methods} & \multirow{2}{*}{Extra} & \multicolumn{3}{c|}{\textit{$AP_{3D|R40|IoU\ge0.7}$}} & \multicolumn{3}{c}{\textit{$AP_{BEV|R40|IoU\ge0.7}$}}\\
 & & Easy & Mod. & Hard & Easy & Mod. & Hard \\
\midrule

Kinematic3D \cite{kinematic3d} & Multi-frames & 19.76 & 14.10 & 10.47 & 27.83 & 19.72 & 15.10 \\
    
CaDDN \cite{caddn} & Lidar & 23.57 & 16.31 & 13.84 & - & - & - \\

\midrule
    
MonoDIS \cite{monodis} & \multirow{8}{*}{None} & 11.06 & 7.60 & 6.37 & 18.45 & 12.58 & 10.66 \\
    
M3D-RPN \cite{m3drpn} & & 14.53 & 11.07 & 8.65 & 20.85 & 15.62 & 11.88 \\
    
MonoPair \cite{monopair} & & 16.28 & 12.30 & 10.42 & 24.12 & 18.17 & 15.76\\
    
MonoDLE \cite{monodle} & & 17.45 & 13.66 & 11.68 & 24.97 & 19.33 & 17.01\\
    
MonoRCNN \cite{monorcnn} & & 16.61 & 13.19 & 10.65 & 25.29 & 19.22 & 15.30\\
    
MonoGeo \cite{monogeo} & & 18.45 & 14.48 & 12.87 & 27.15 & 21.17 & 18.35 \\
    
MonoFlex \cite{monoflex} & & \textcolor{blue}{23.64} & \textcolor{blue}{17.51} & \textcolor{blue}{14.83} & - & - & - \\
    
GUPNet \cite{gupnet} & & 22.76 & 16.46 & 13.72 & \textcolor{blue}{31.07} & \textcolor{blue}{22.94} & \textcolor{blue}{19.75} \\

\midrule

\textbf{MonoCon (Ours)} & None & \textbf{26.33} & \textbf{19.01} & \textbf{15.98} & \textbf{34.65} & \textbf{25.39} & \textbf{21.93}\\
    
\bottomrule
\end{tabular}
\end{center}
\caption{Quantitative performance of the \textbf{Car} category on the KITTI \textit{validation} set. Method are ranked by moderate settings based on 3D detection performance following KITTI leaderboard within each group. We highlight the best results in \textbf{bold} and the second place in \textcolor{blue}{blue}.}
\label{Tab:val}
\end{table*}

\begin{figure*}[ht]
    \centering
    \includegraphics[width=0.48\textwidth]{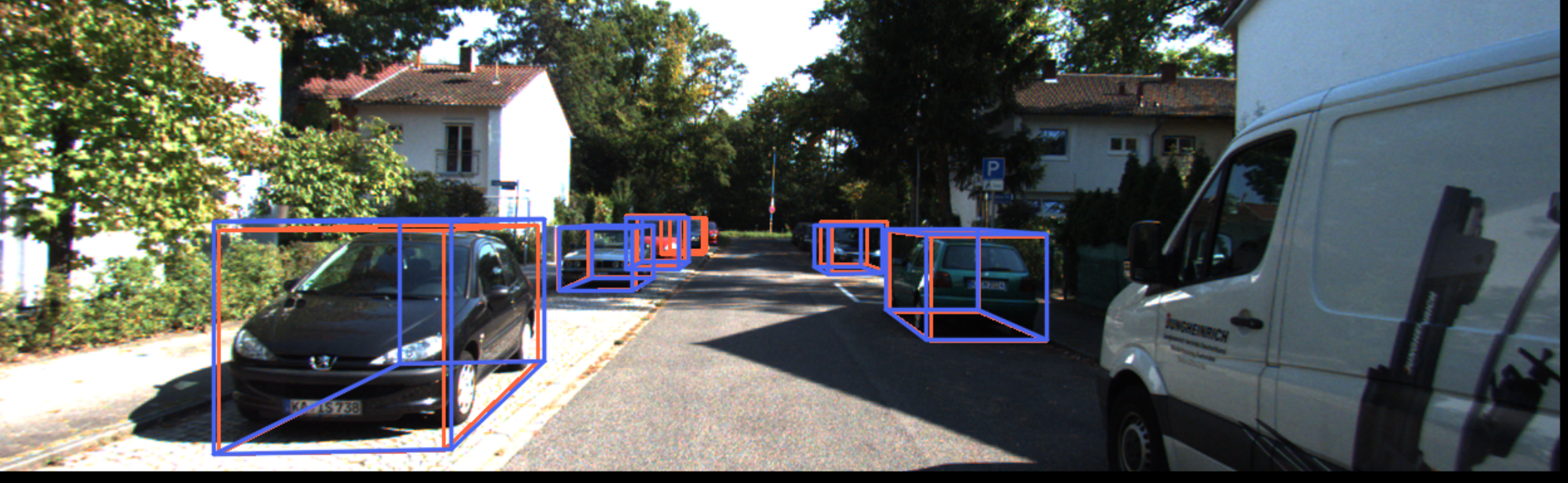}
    \includegraphics[width=0.48\textwidth]{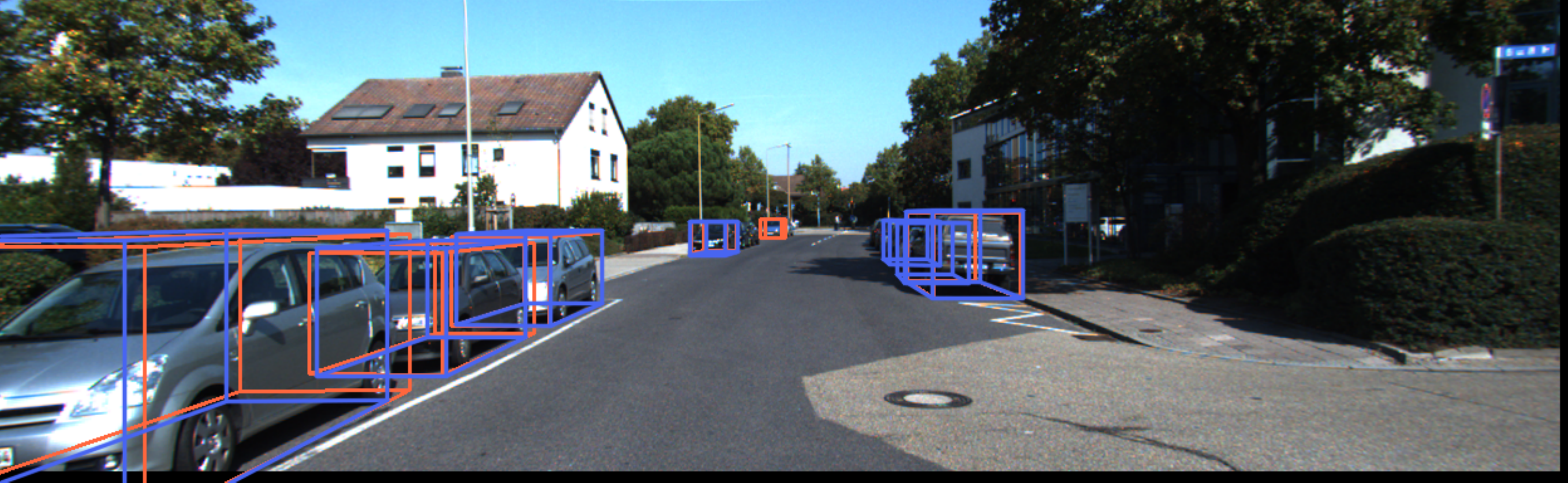} \\
    \includegraphics[width=0.48\textwidth]{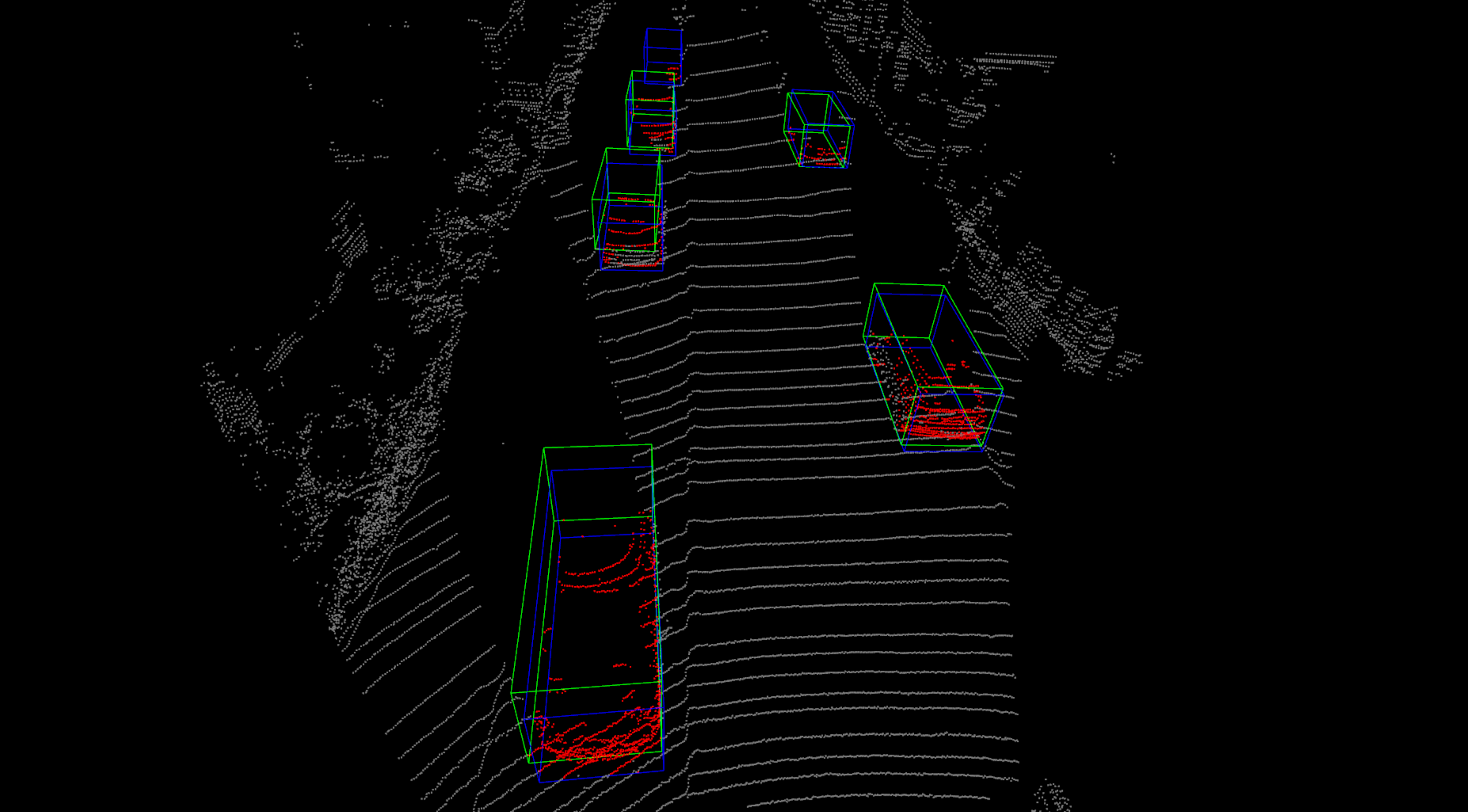}
    \includegraphics[width=0.48\textwidth]{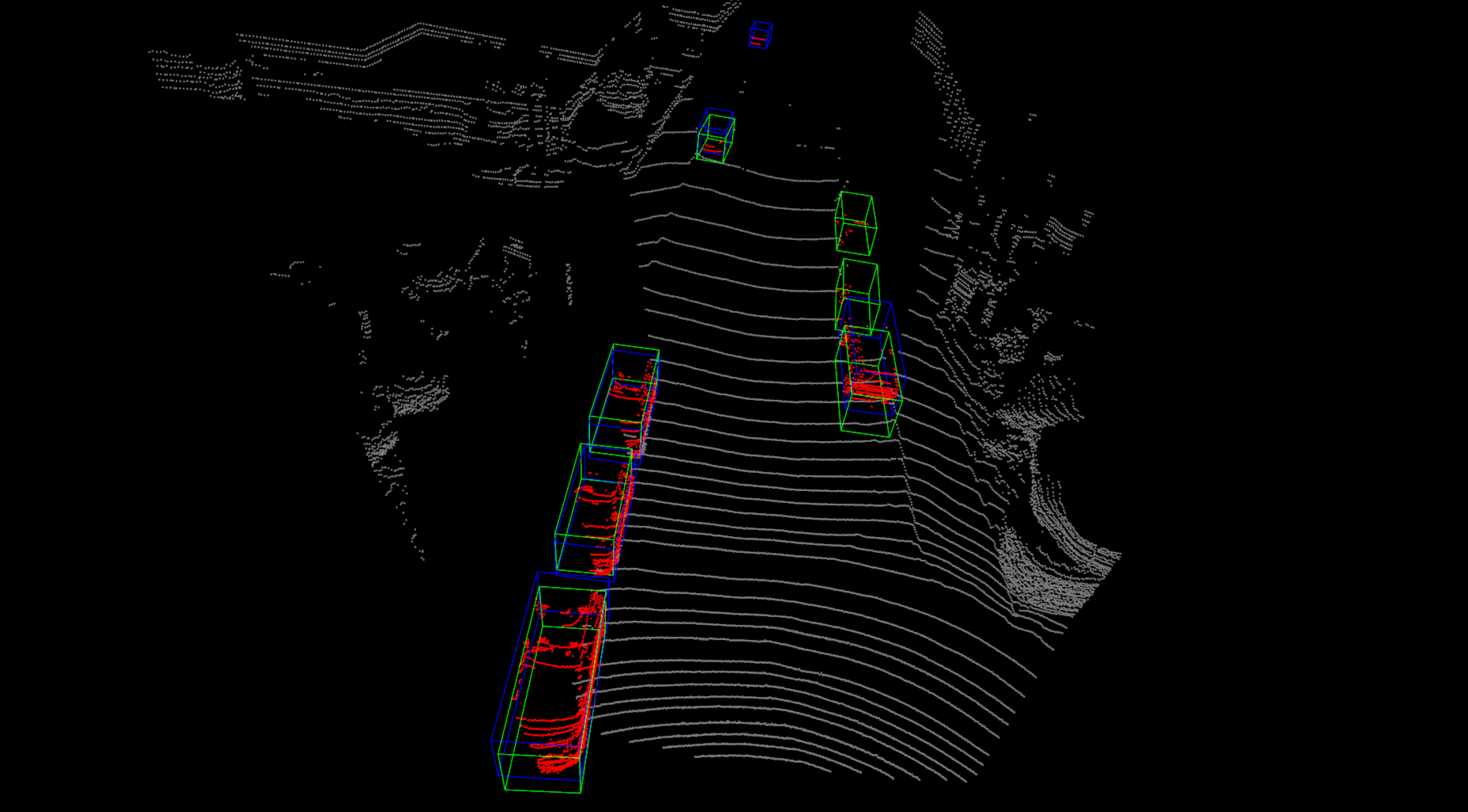} \\
    \includegraphics[width=0.48\textwidth]{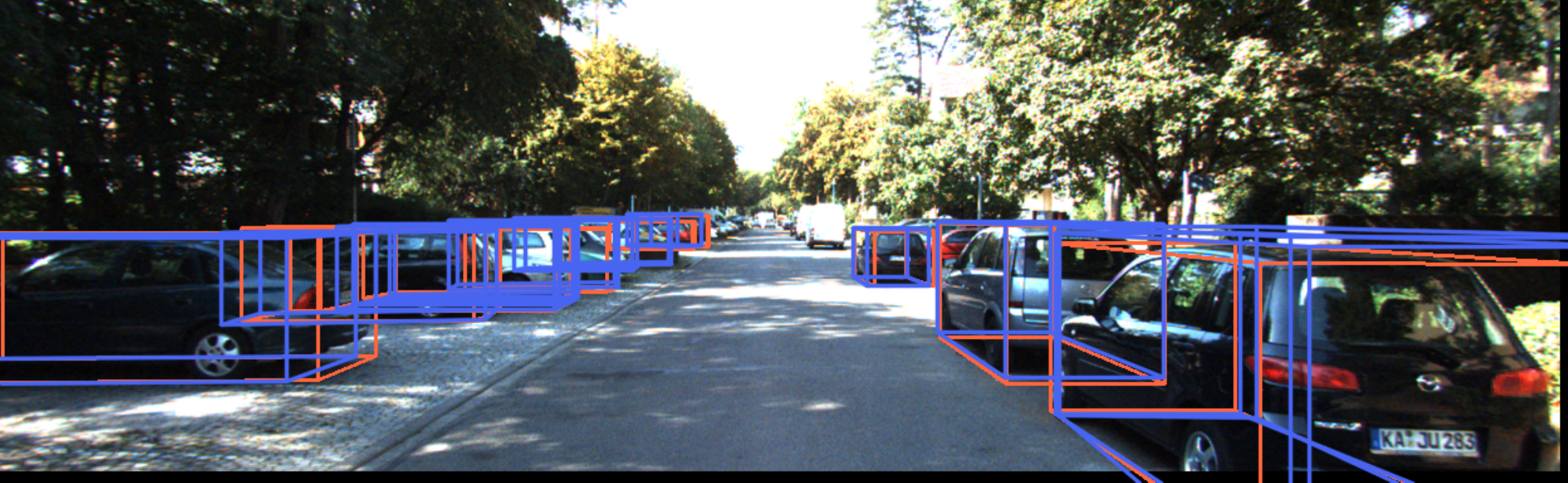}
    \includegraphics[width=0.48\textwidth]{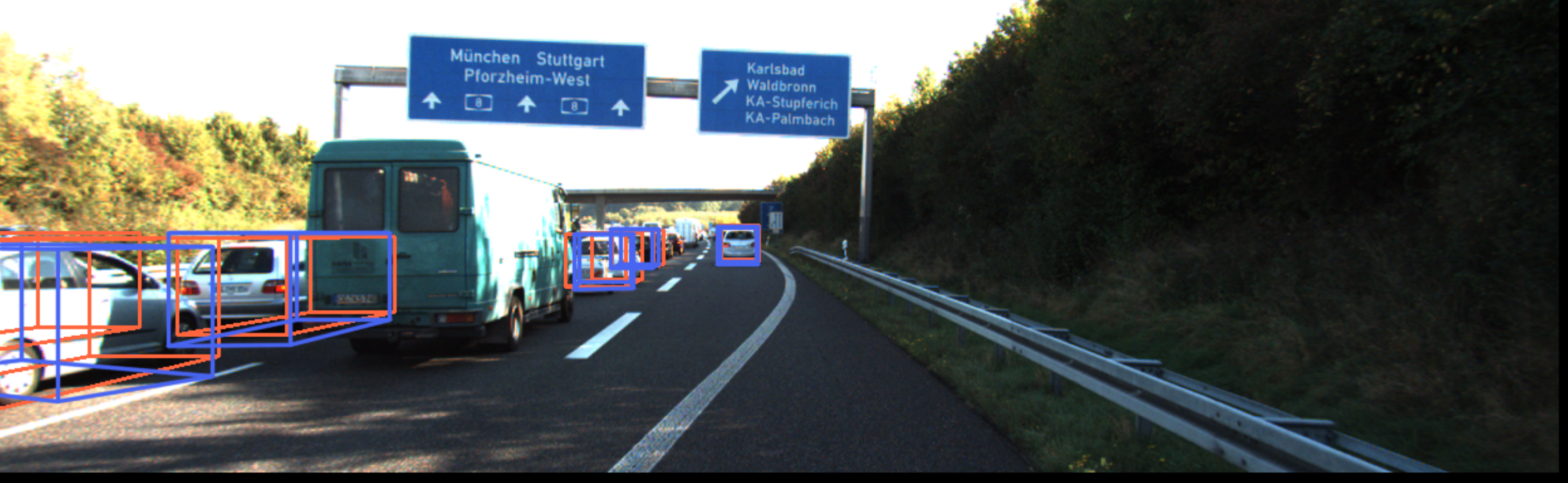} \\
    \includegraphics[width=0.48\textwidth]{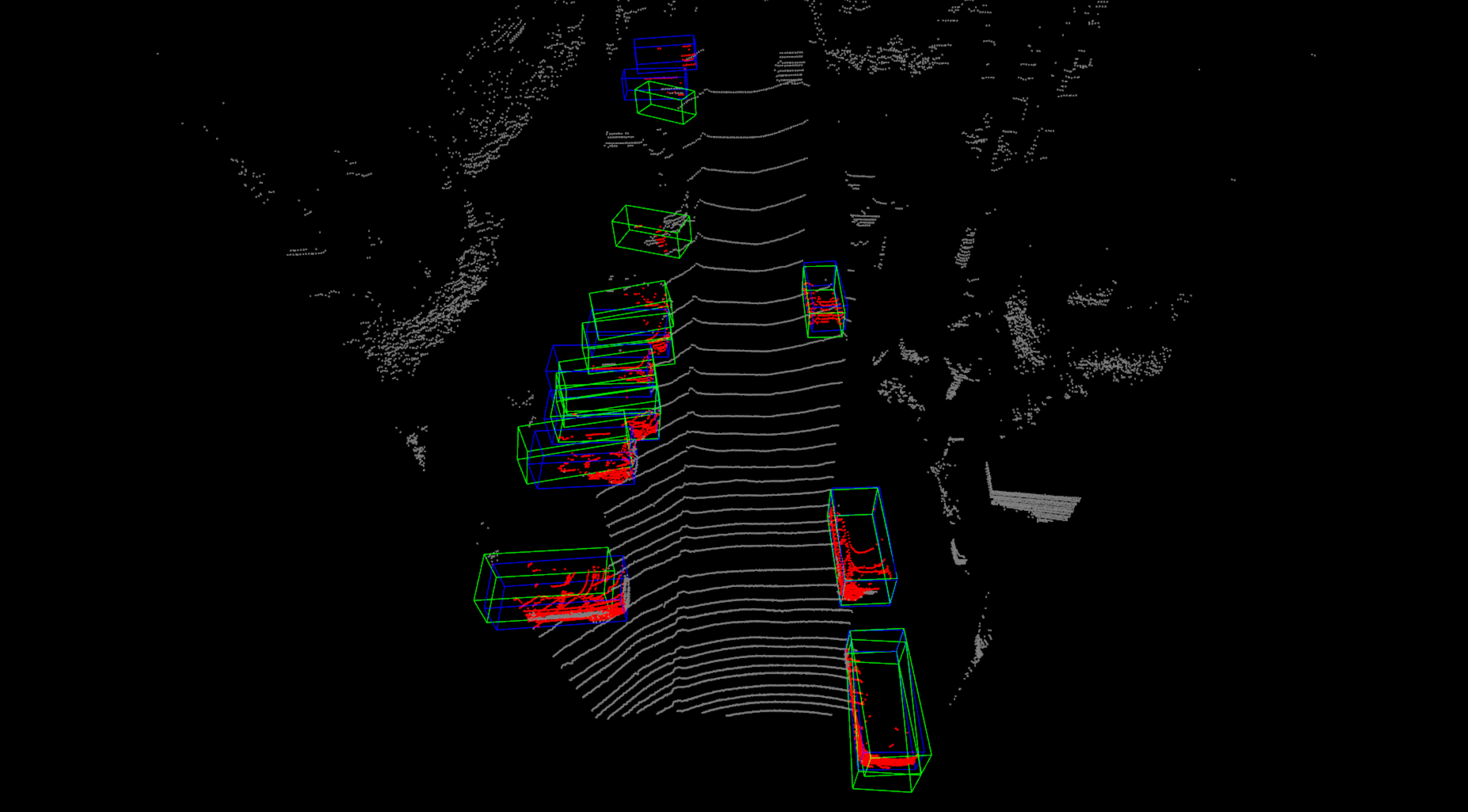}
    \includegraphics[width=0.48\textwidth]{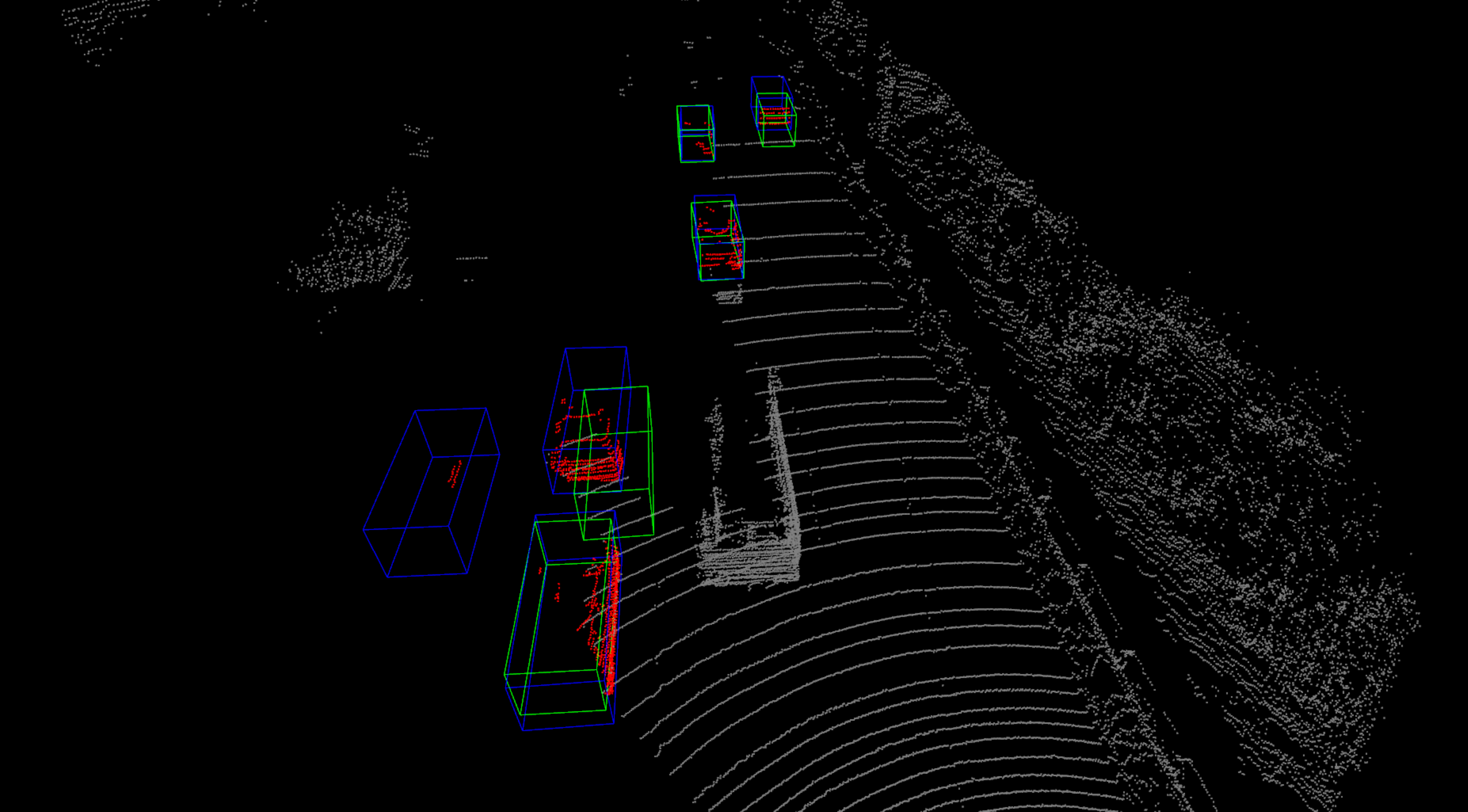} \\
    \includegraphics[width=0.48\textwidth]{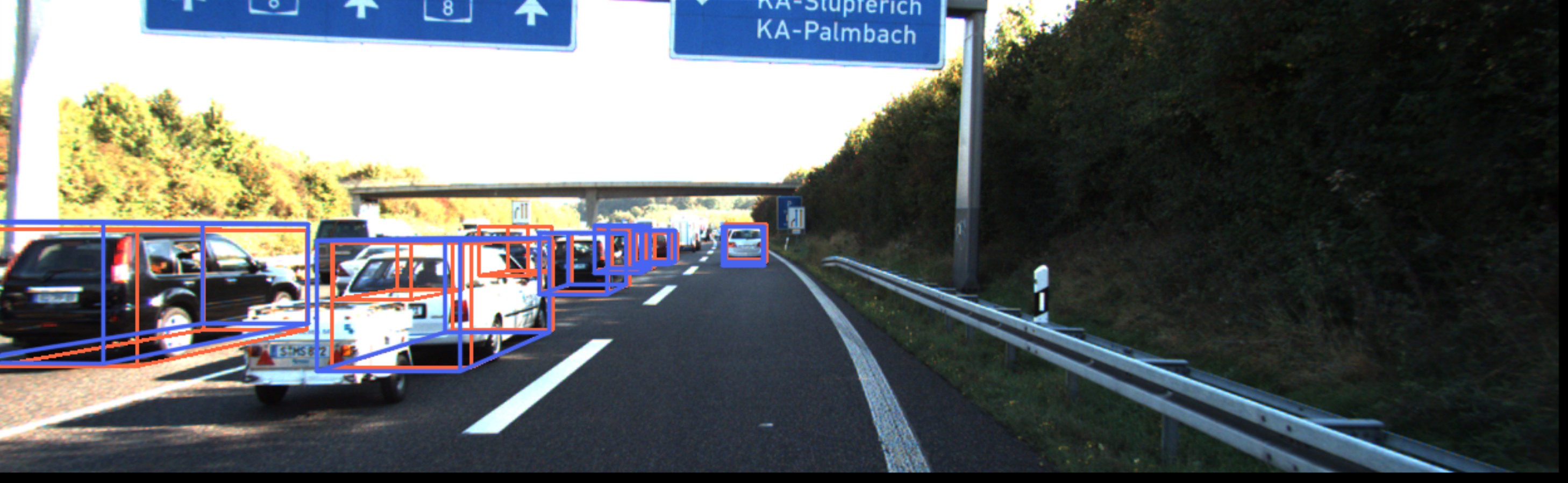}
    \includegraphics[width=0.48\textwidth]{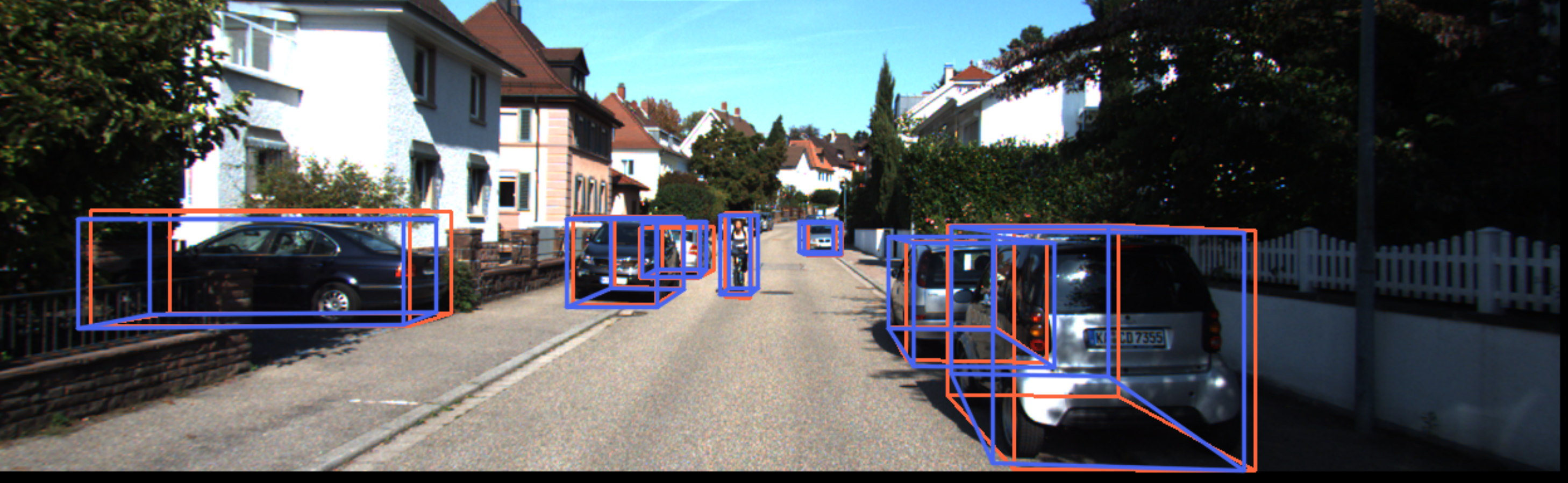} \\
    \includegraphics[width=0.48\textwidth]{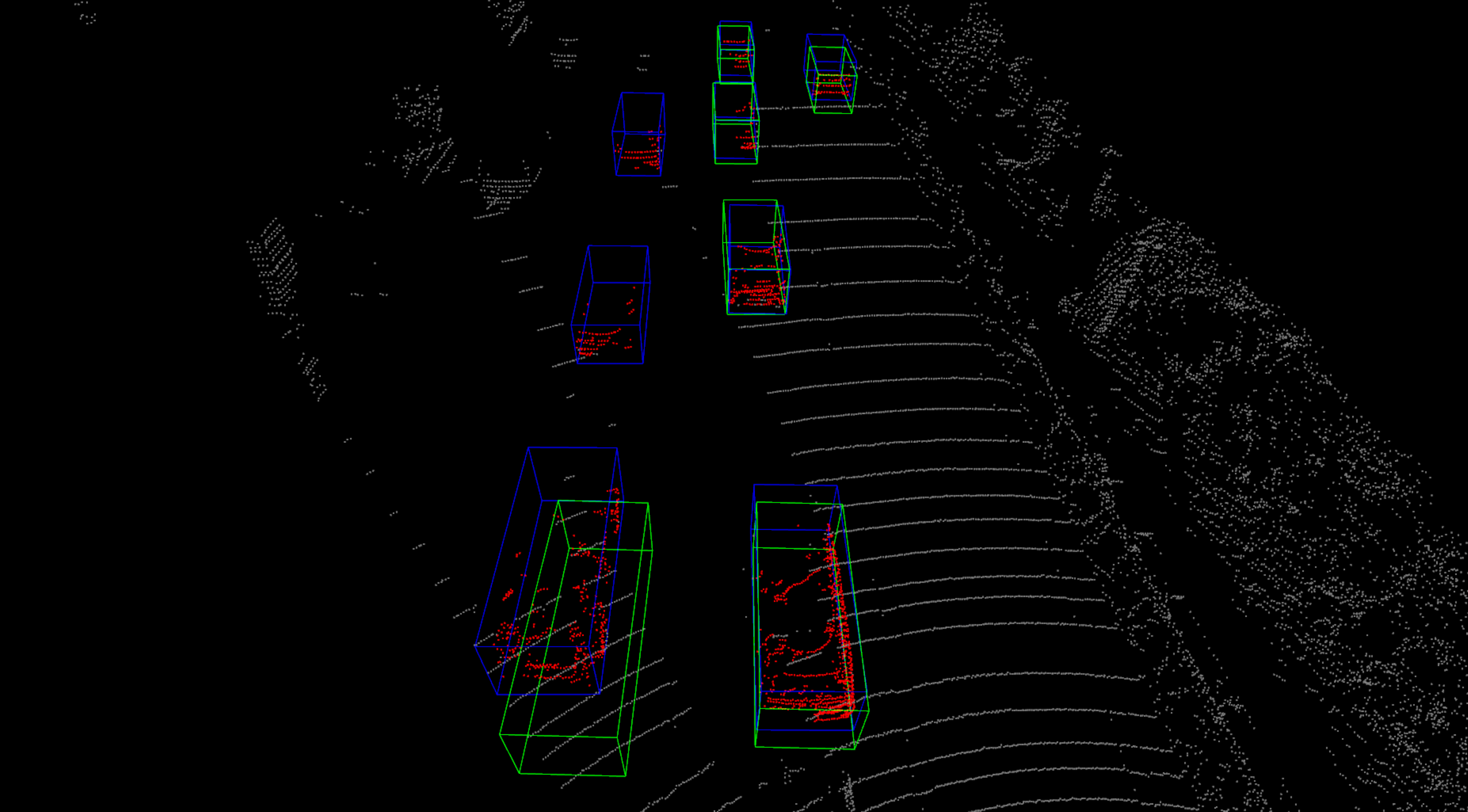}
    \includegraphics[width=0.48\textwidth]{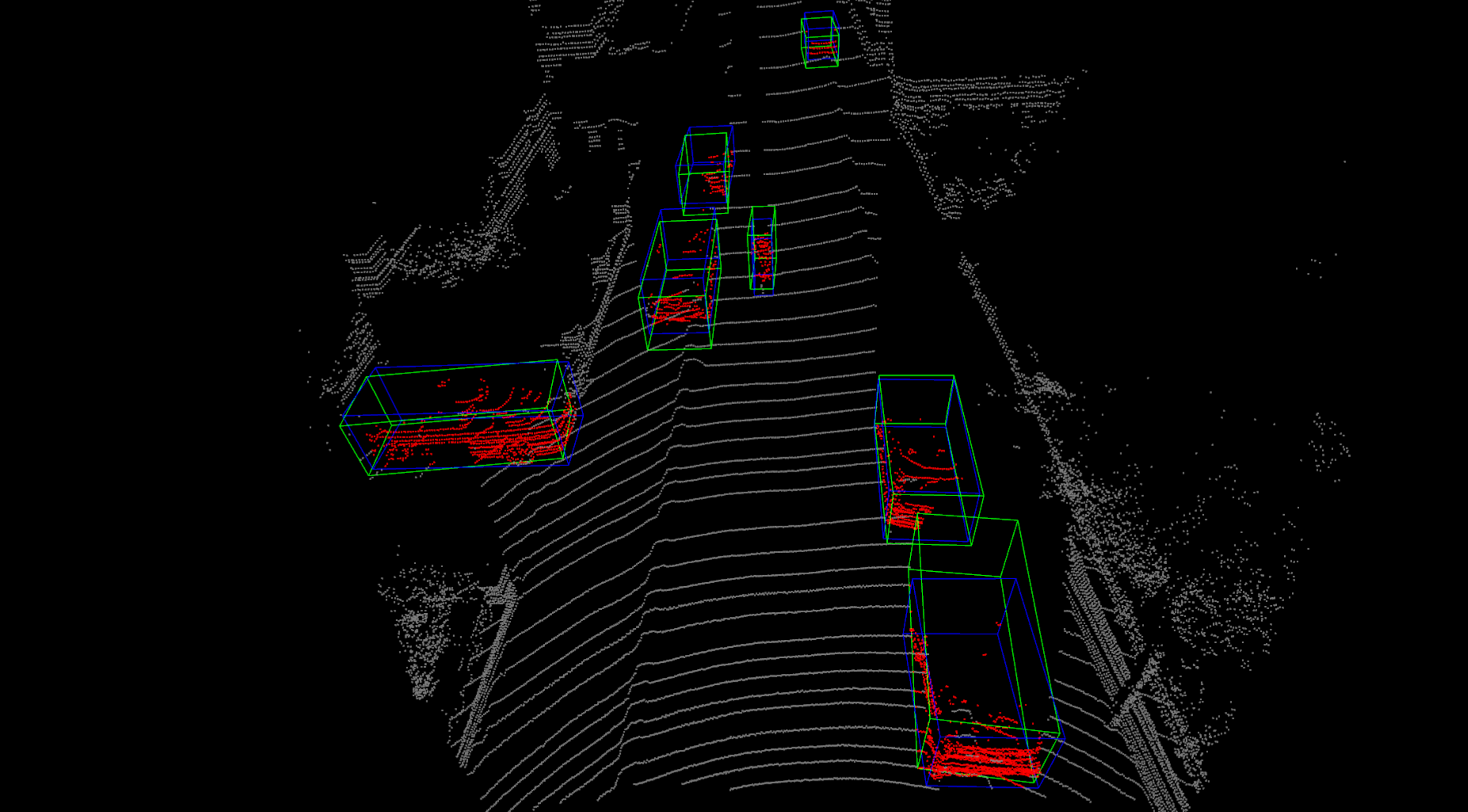} \\
    \caption{Qualitative results of our MonoCon on KITTI \textit{validation} set \cite{mono3d}. In the front view image, our prediction result is shown in \textcolor{blue}{blue}, while the ground truth is shown in \textcolor{orange}{orange}. In the lidar view image, our prediction result is shown in \textcolor{green}{green}. The ground truth 3D box is shown in \textcolor{blue}{blue}.}
    \label{fig:qual_val}
\end{figure*}

\begin{figure*}
    \centering
    \includegraphics[width=0.48\textwidth]{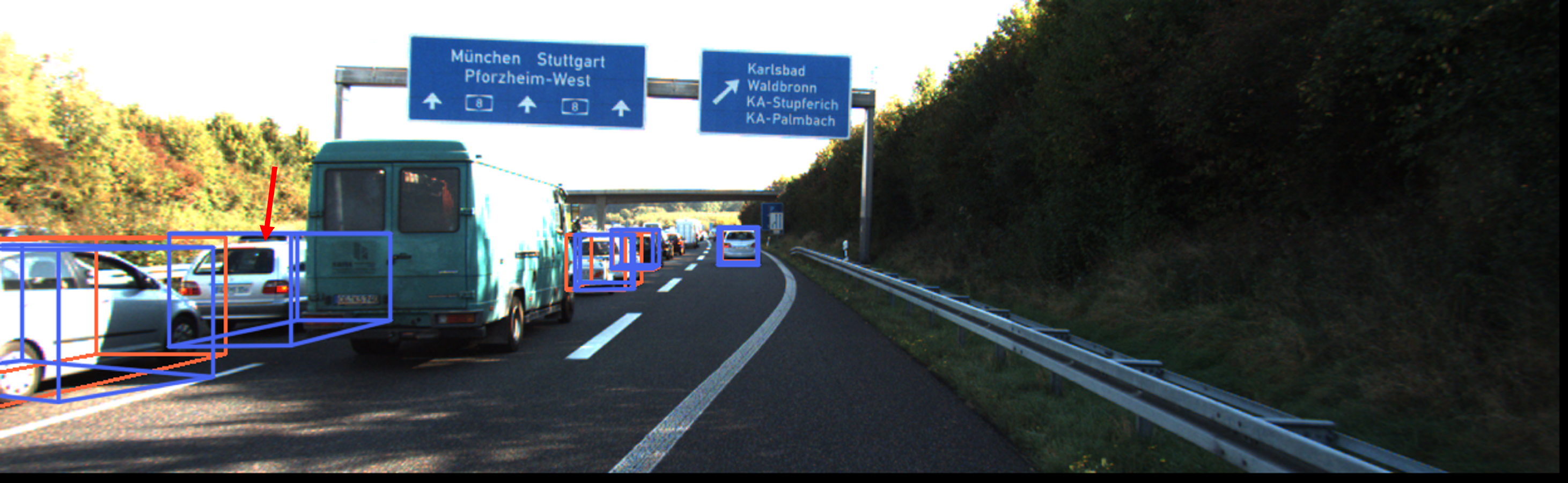}
    \includegraphics[width=0.48\textwidth]{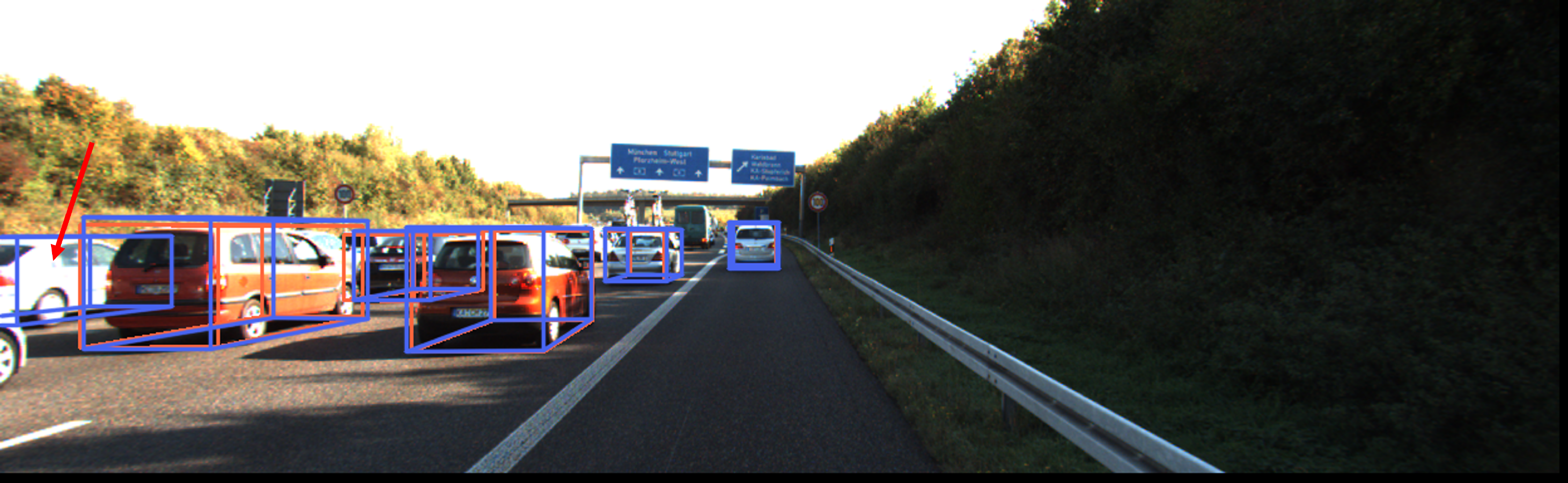} \\
    \includegraphics[width=0.48\textwidth]{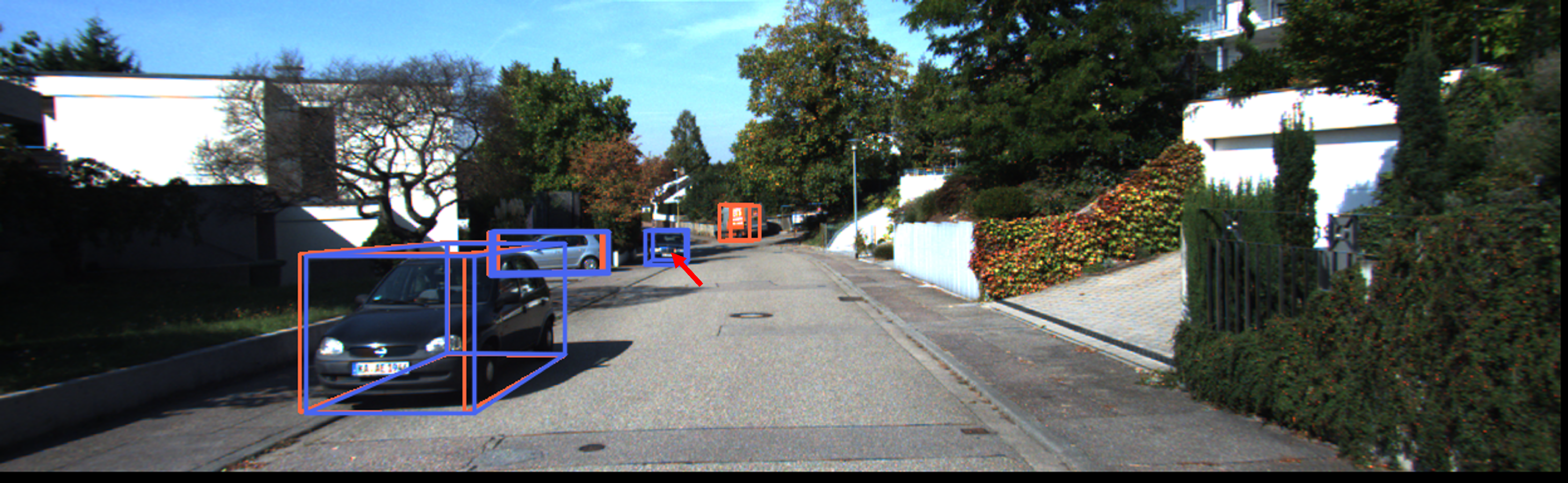}
    \includegraphics[width=0.48\textwidth]{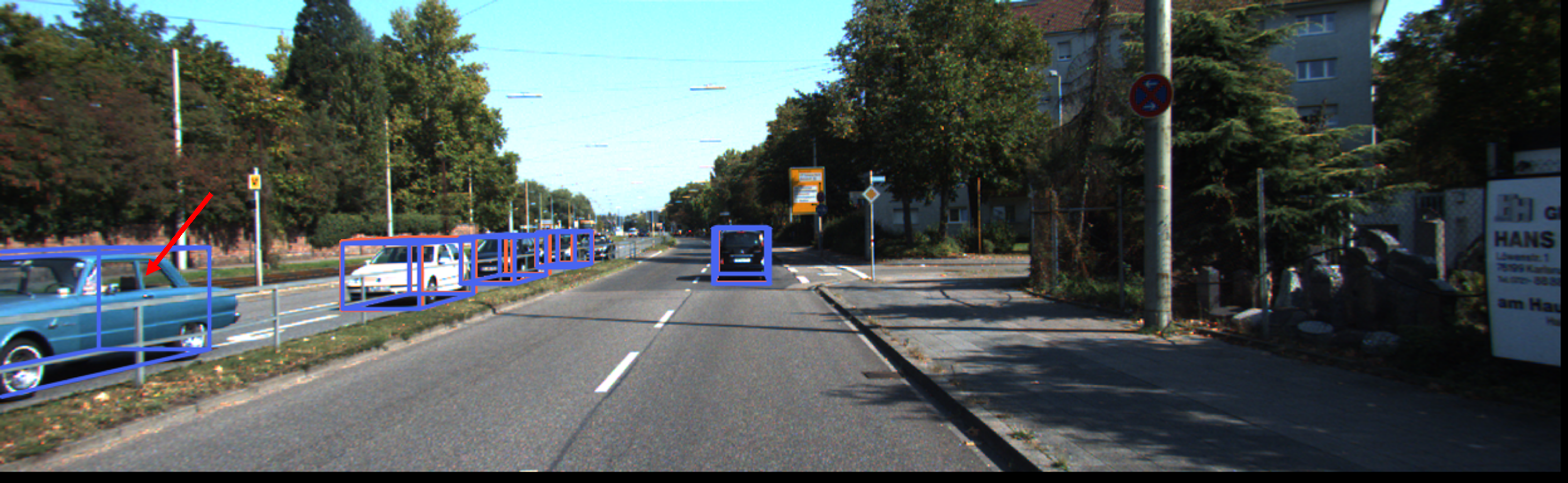} \\
    \includegraphics[width=0.48\textwidth]{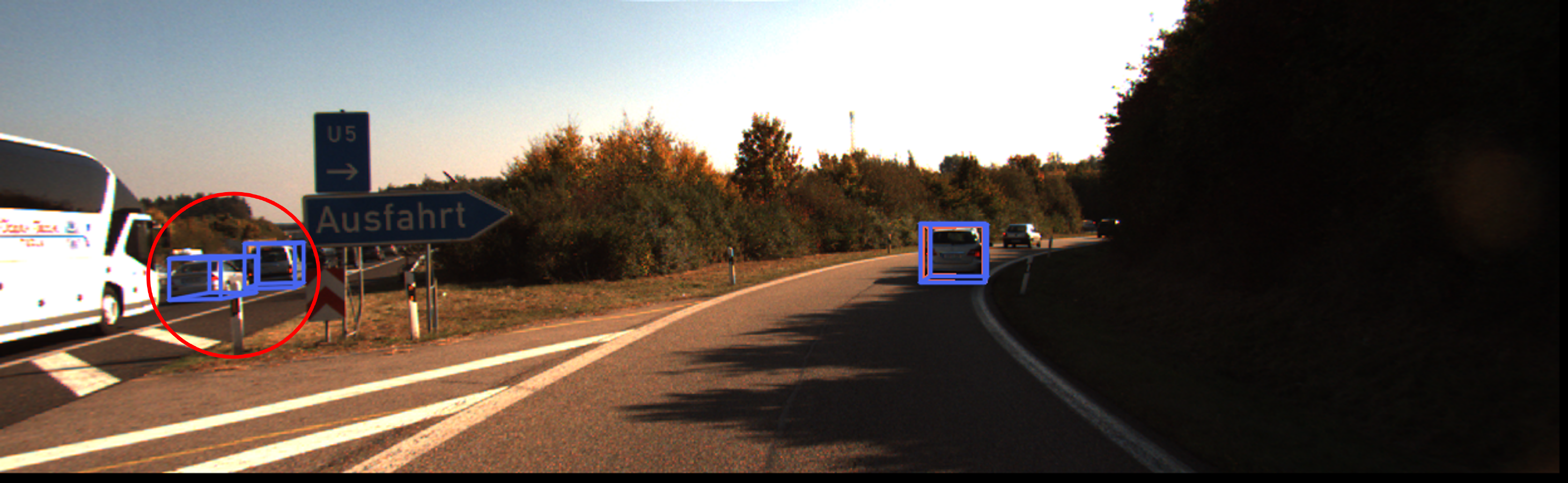}
    \includegraphics[width=0.48\textwidth]{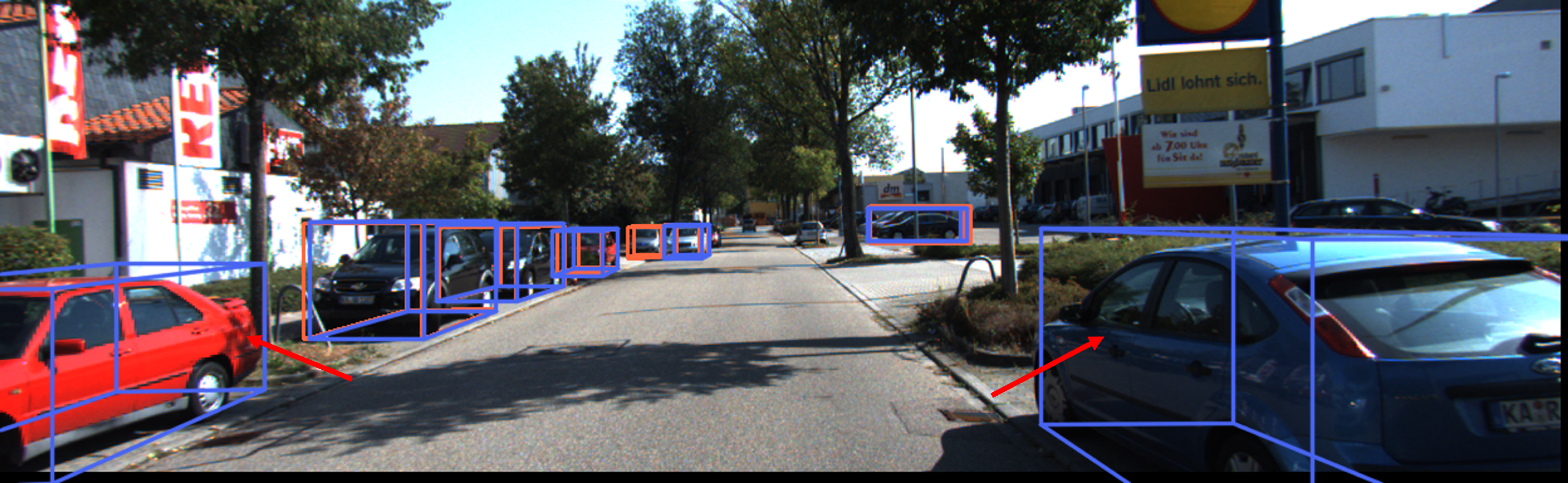} \\

    \caption{Comparison of our MonoCon with enhanced MonoDLE* on KITTI \textit{validation} set \cite{mono3d}. Our prediction result is shown in \textcolor{blue}{blue}. The MonoDLE's prediction is shown in \textcolor{orange}{orange}.}
    \label{fig:vs_monodle}
\end{figure*}

\paragraph{Qualitative Results:} We show more qualitative results on the validation set in Fig.~\ref{fig:qual_val}. It shows that our MonoCon method achieves high localization performance on close\&mid-distance, not heavily occluded instances. For the instances that are heavily occluded and far away from the viewpoint, it would be still challenging for both our MonoCon and other monocular 3D object detectors~\cite{monodle, monoef, monoflex, gupnet}.
Furthermore, we compare our MonoCon with our implemented enhanced MonoDLE ~\cite{monodle} in Fig.~\ref{fig:vs_monodle} to demonstrate the advantages of our MonoCon. The results shows that our MonoCon achieves better performance on truncated, occluded and far instances than the best-performing prior art.

\section{More Qualitative Results on KITTI Test Set}

\begin{figure*}
    \centering
    \includegraphics[width=0.48\textwidth]{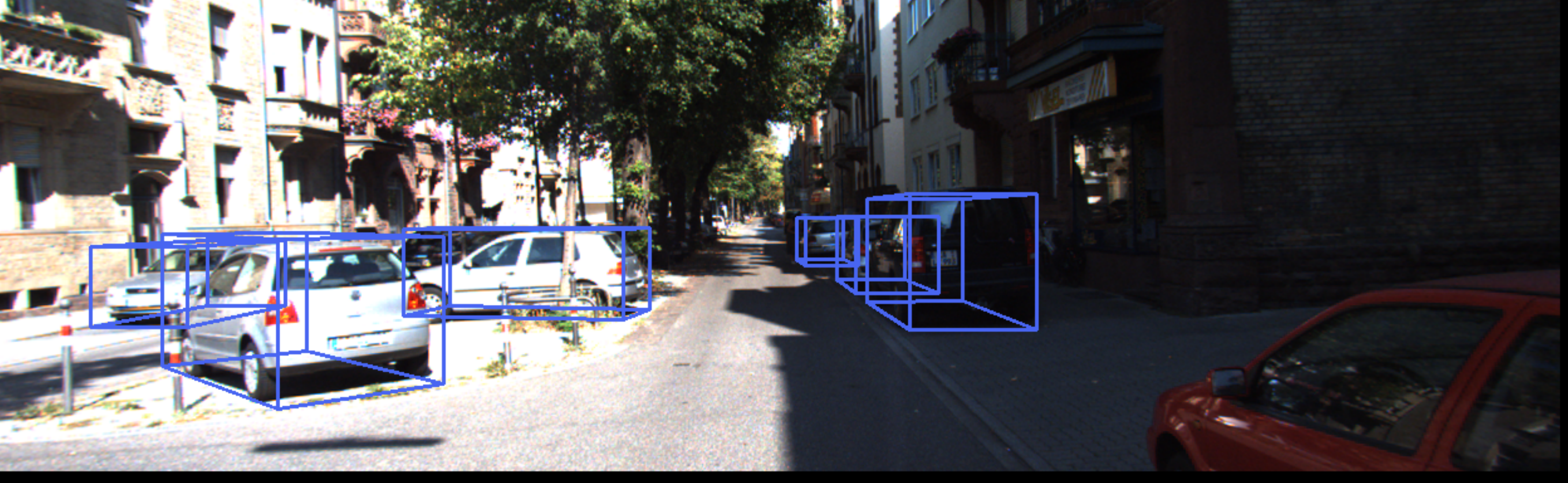}
    \includegraphics[width=0.48\textwidth]{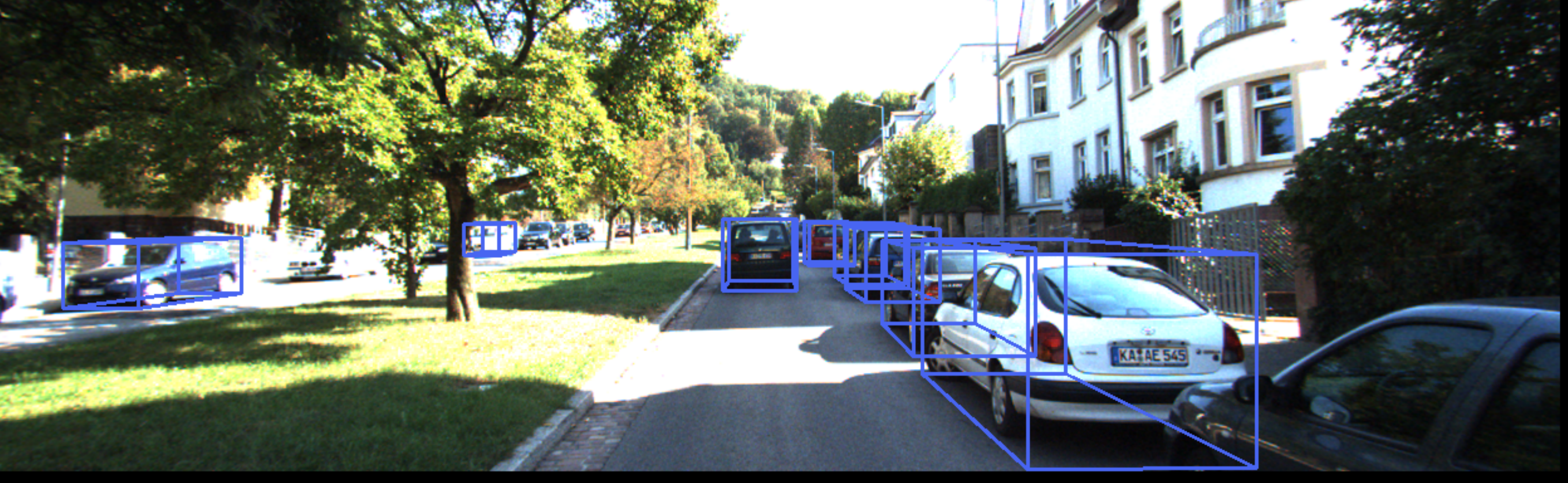} \\
    \includegraphics[width=0.48\textwidth]{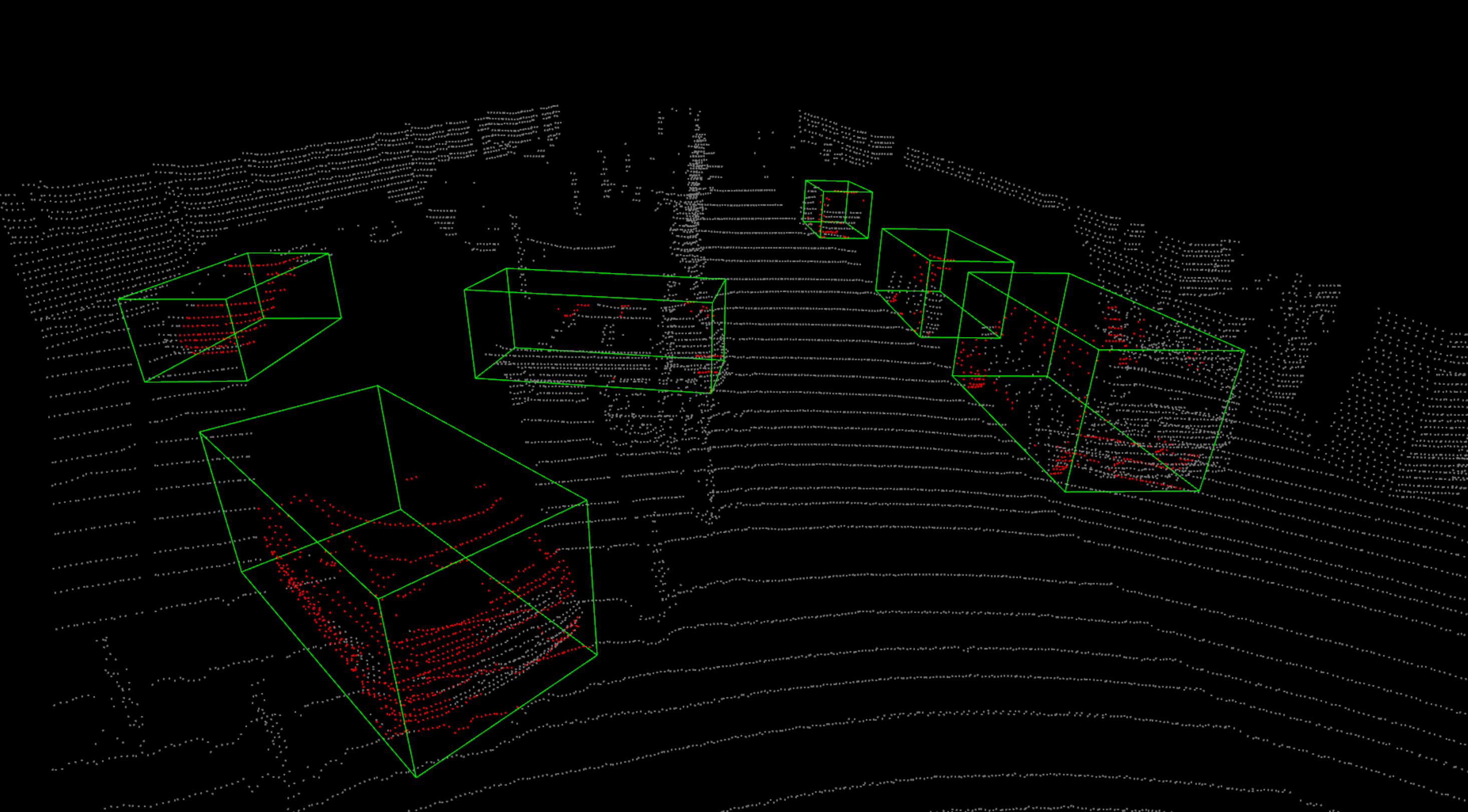}
    \includegraphics[width=0.48\textwidth]{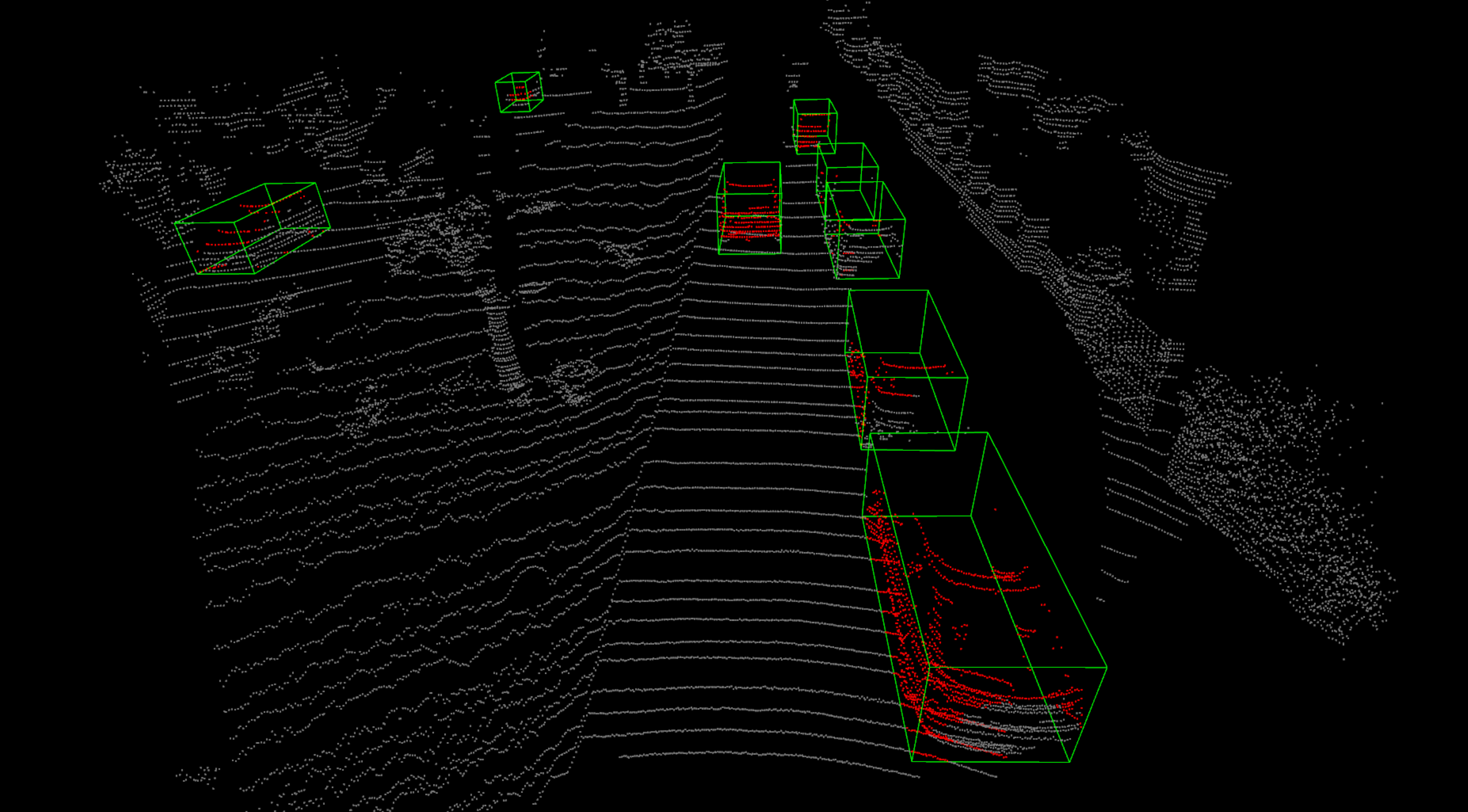} \\
    \includegraphics[width=0.48\textwidth]{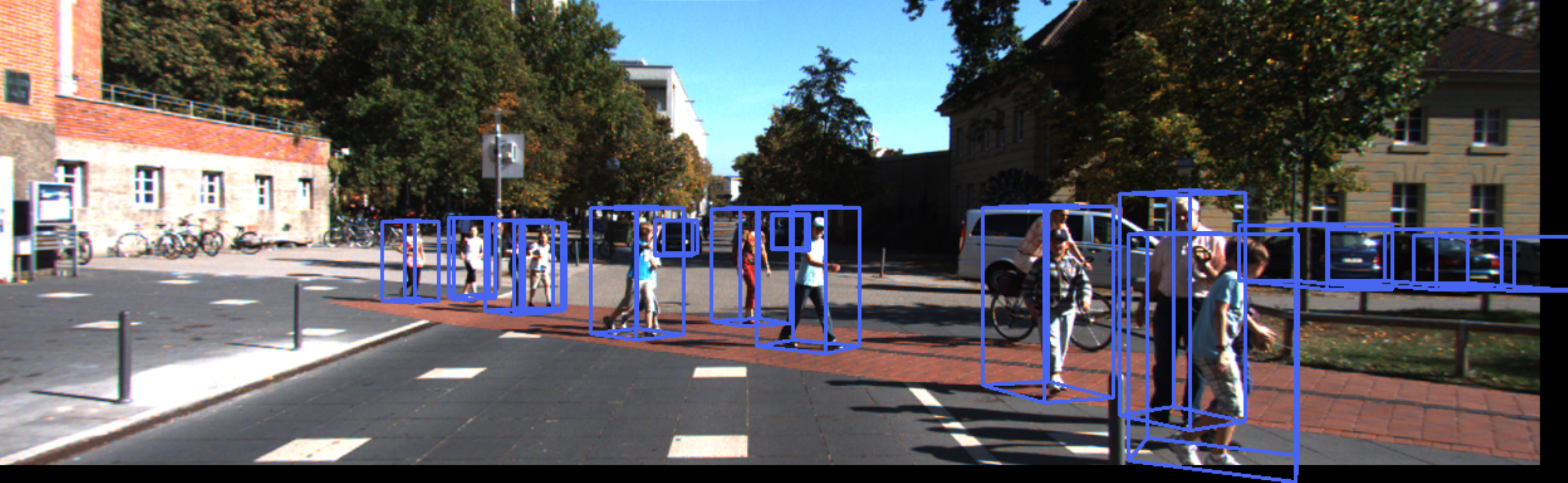}
    \includegraphics[width=0.48\textwidth]{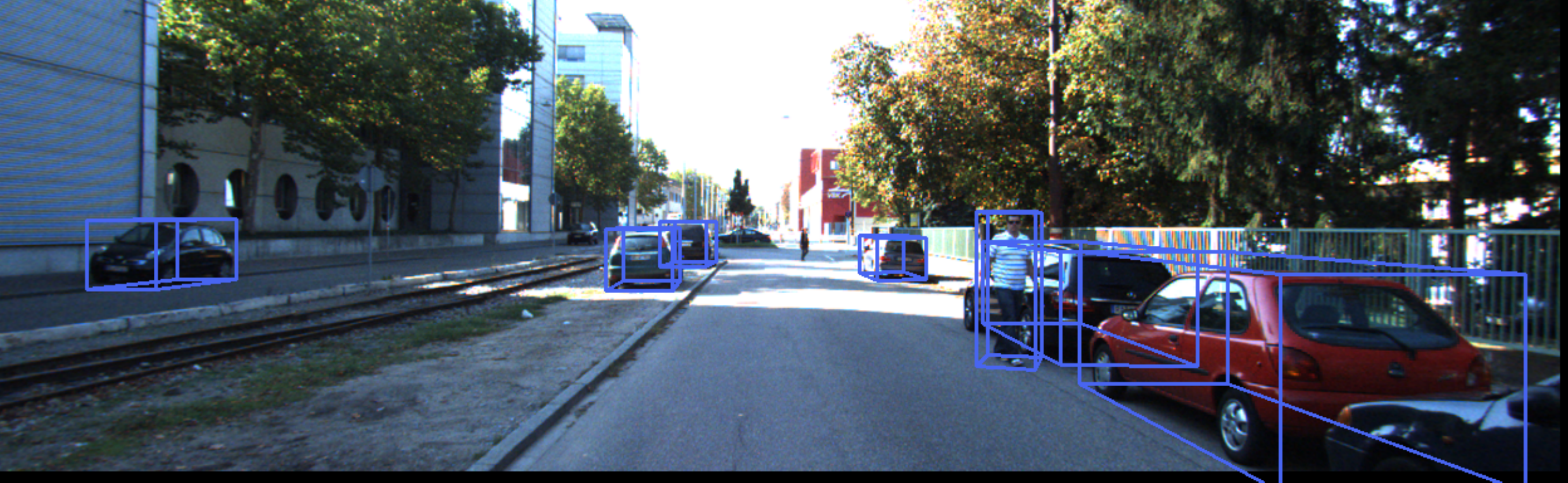} \\
    \includegraphics[width=0.48\textwidth]{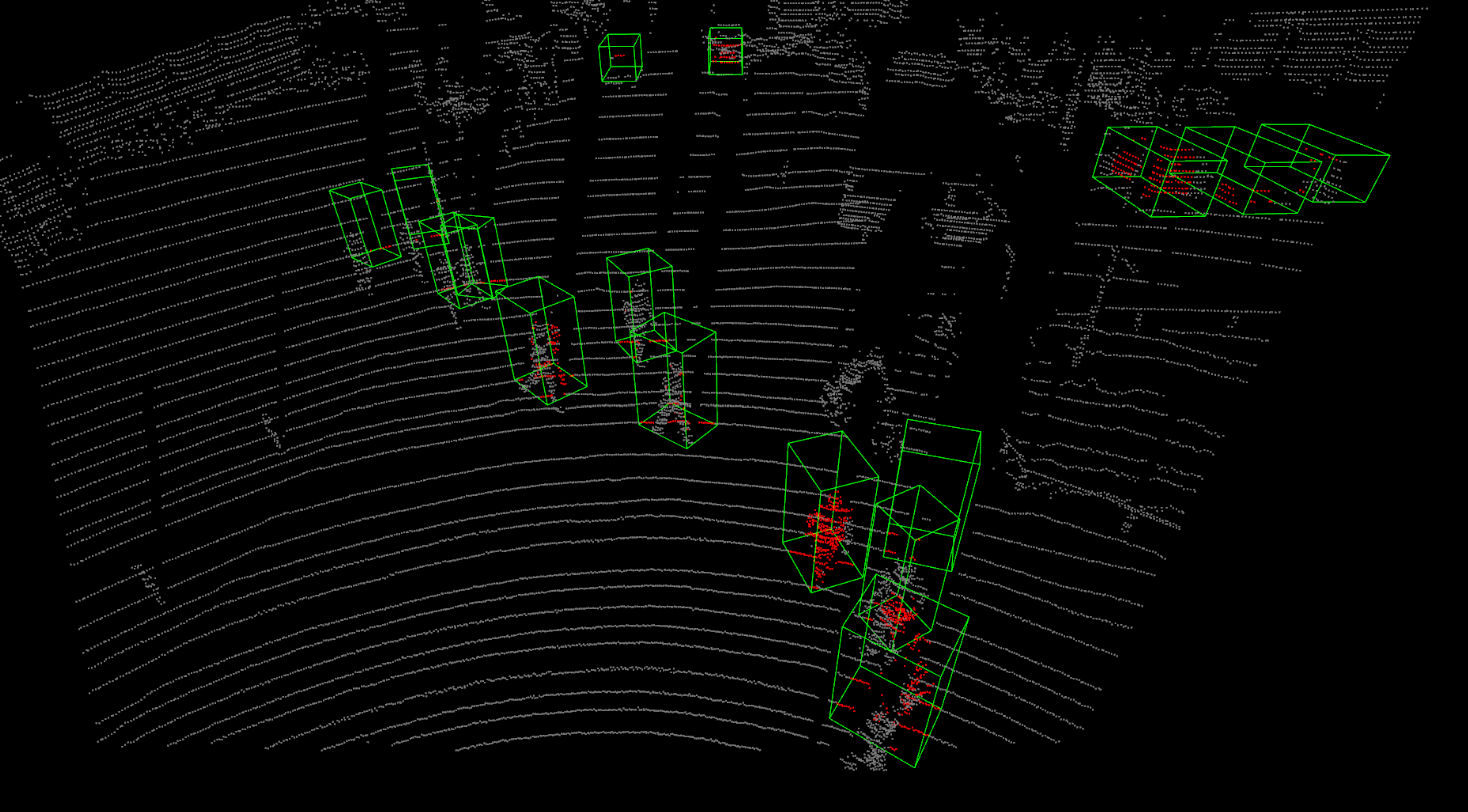}
    \includegraphics[width=0.48\textwidth]{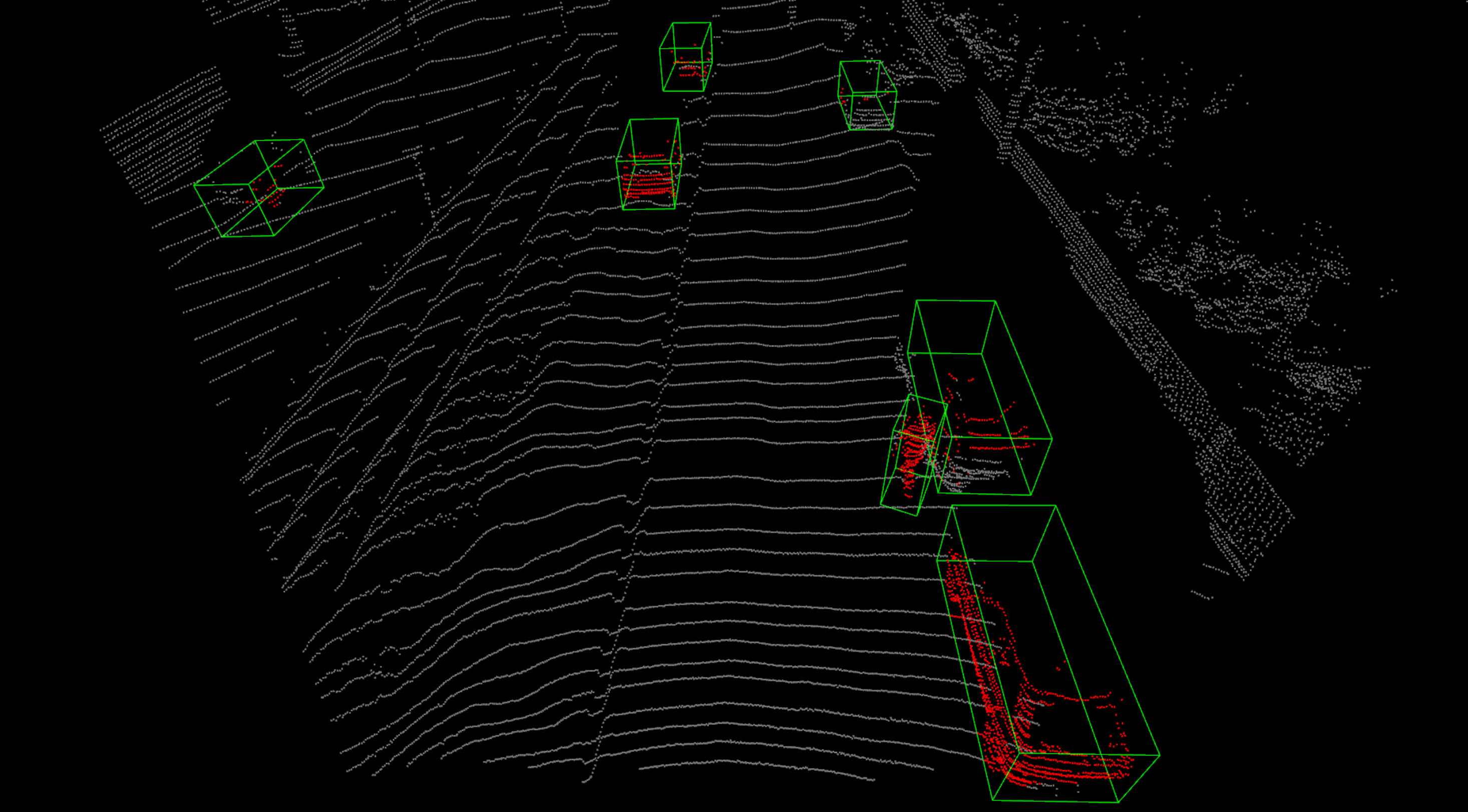} \\
    \includegraphics[width=0.48\textwidth]{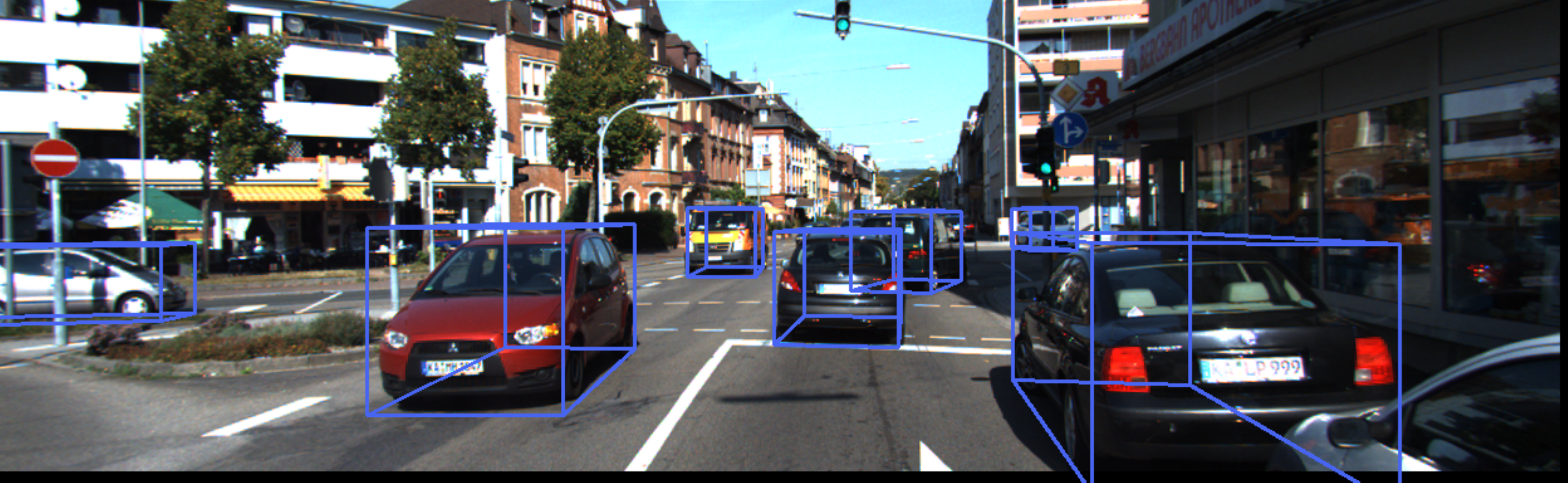}
    \includegraphics[width=0.48\textwidth]{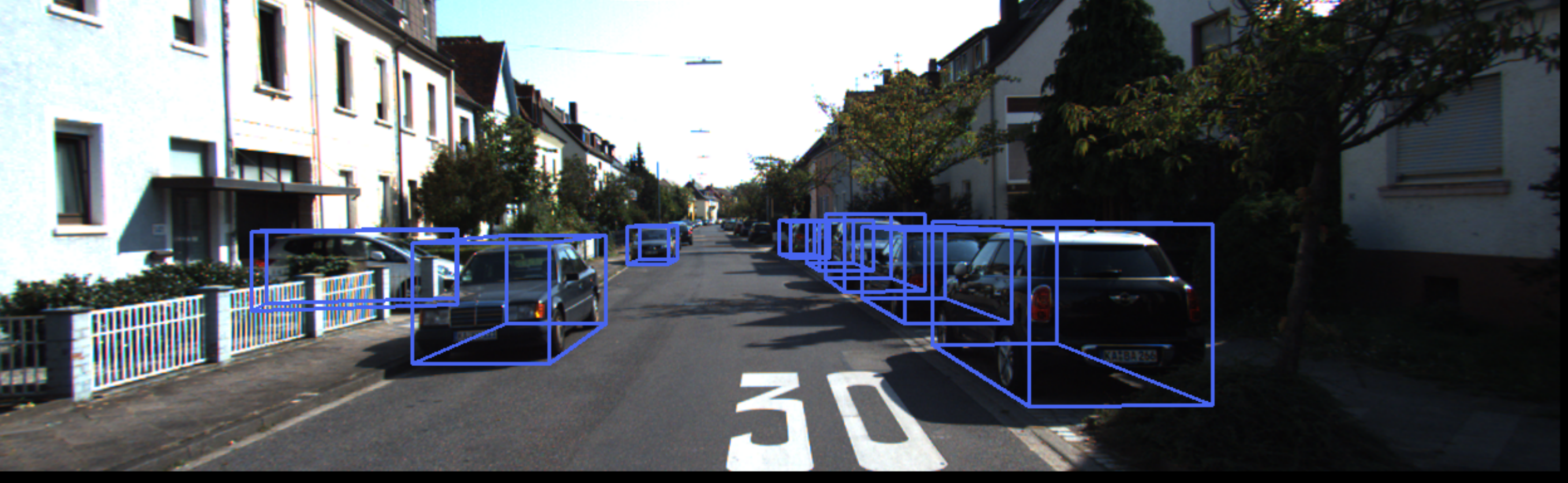} \\
    \includegraphics[width=0.48\textwidth]{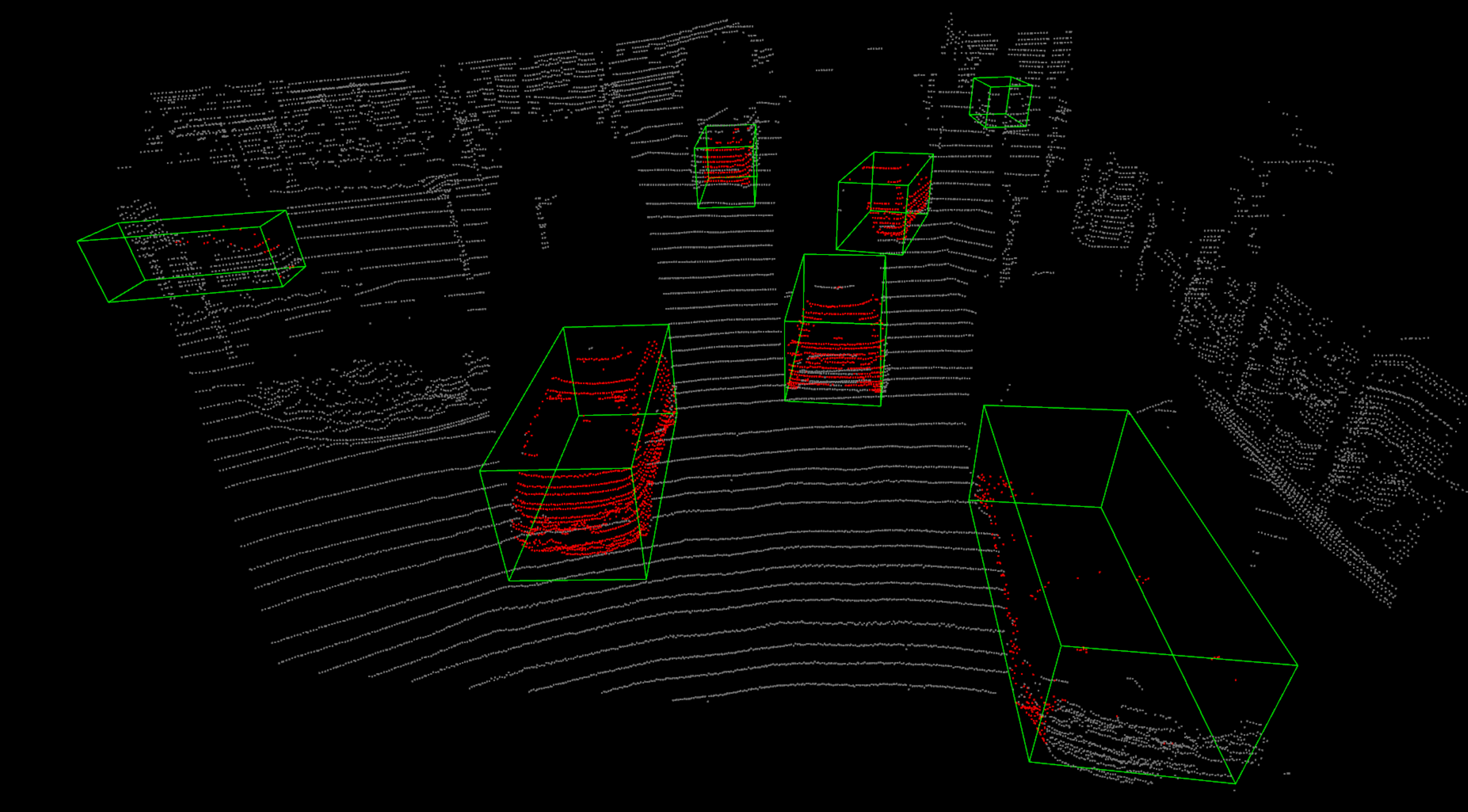}
    \includegraphics[width=0.48\textwidth]{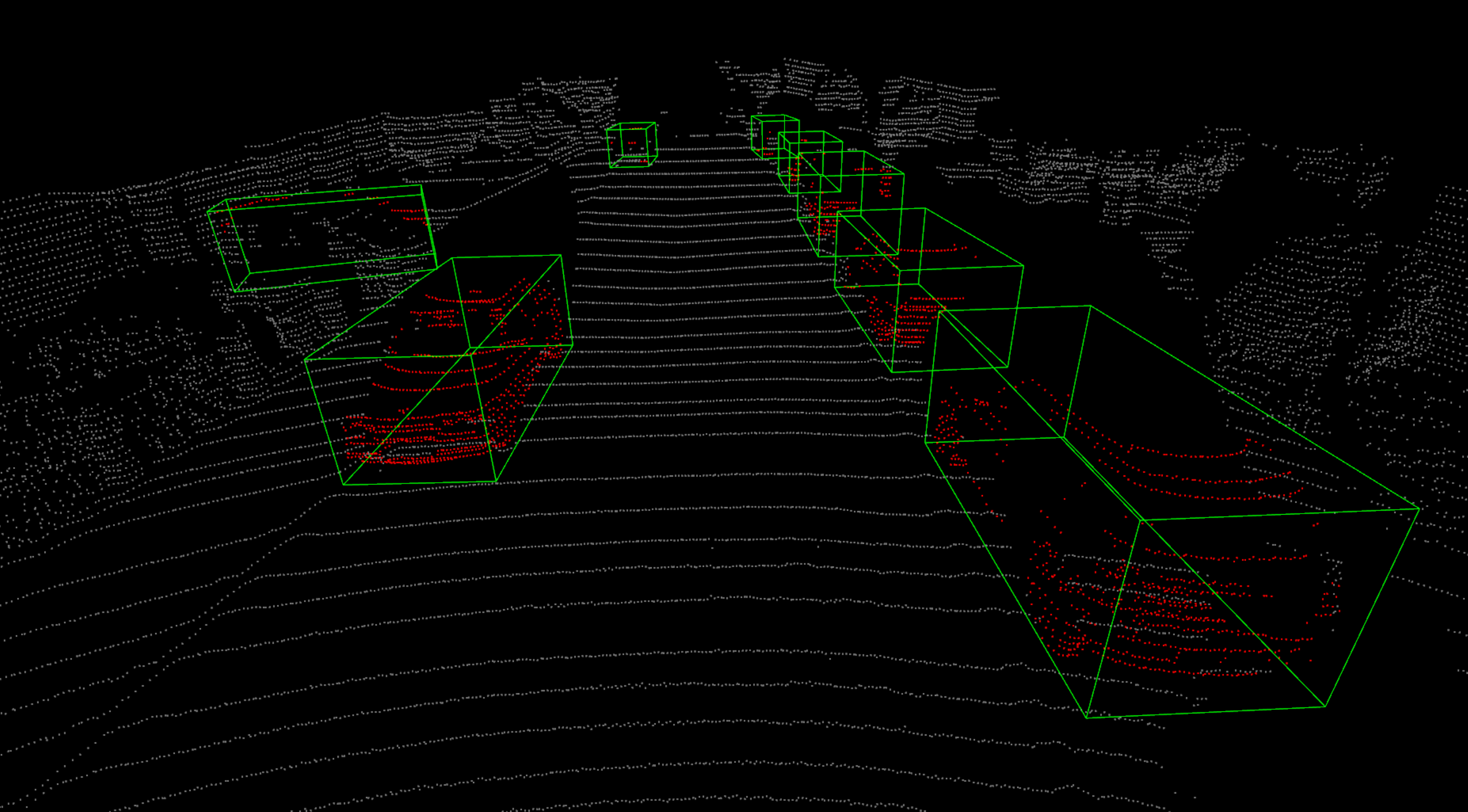} \\

    \caption{Qualitative results of our MonoCon on KITTI \textit{test} set. In the front view image, our prediction result is shown in \textcolor{blue}{blue}. In the lidar view image, our prediction result is shown in \textcolor{green}{green}.}
    \label{fig:qual_test}
\end{figure*}

We present our MonoCon's qualitative 3D detection performance in Fig.~\ref{fig:qual_test}. Qualitative result shows that our model achieves high 3D detection precision, i.e. our model's predicted 3D box encompasses instances' point cloud well. Besides, our model even predicts two close objects (i.e. a person next to a car), which is usually hard for prior arts.

\end{document}